\documentclass[sigconf]{acmart}

\AtBeginDocument{%
  \providecommand\BibTeX{{%
    \normalfont B\kern-0.5em{\scshape i\kern-0.25em b}\kern-0.8em\TeX}}}

\setcopyright{acmcopyright}
\copyrightyear{2018}
\acmYear{2018}
\acmDOI{10.1145/1122445.1122456}

\acmConference[Woodstock '18]{Woodstock '18: ACM Symposium on Neural
  Gaze Detection}{June 03--05, 2018}{Woodstock, NY}
\acmBooktitle{Woodstock '18: ACM Symposium on Neural Gaze Detection,
  June 03--05, 2018, Woodstock, NY}
\acmPrice{15.00}
\acmISBN{978-1-4503-XXXX-X/18/06}

\usepackage[ruled,vlined, linesnumbered]{algorithm2e}
\usepackage{bm}
\usepackage{dblfloatfix}

\newcommand{\mt}{\top}

\DeclareMathOperator*{\argmax}{arg\,max}

\def \btheta {\mathrm{\boldsymbol{\theta}}}

\def \bx {\mathbf{x}}

\def \cA {\mathcal{A}}
\def \bb {\mathbf{b}}
\def \bI {\mathbf{I}}

\def \cN {\mathcal{N}}

\def \cC {\mathcal{C}}
\def \cI {\mathcal{I}}
\def \cM {\mathcal{M}}
\def \cG {\mathcal{G}}

\def \bbP {\mathbb{P}}
\def \bbE {\mathbb{E}}
\def \bbR {\mathbb{R}}

\def \cH {\mathcal{H}}
\def \cD {\mathcal{D}}
\def \cN {\mathcal{N}}
\def \cG {\mathcal{G}}
\def \cU {\mathcal{U}}
\def \cZ {\mathcal{Z}}
\def \cL {\mathcal{L}}

\begin{document}

\title[Collaborative Dynamic Bandit]{When and Whom to Collaborate with in a Changing Environment: A Collaborative Dynamic Bandit Solution}

\author{Chuanhao Li}
\affiliation{%
  \institution{Department of Computer Science University of Virginia}
}
\email{cl5ev@virginia.edu}

\author{Qingyun Wu}
\affiliation{%
  \institution{Department of Computer Science University of Virginia}
}
\email{qw2ky@virginia.edu}

\author{Hongning Wang}
\affiliation{%
  \institution{Department of Computer Science University of Virginia}
}
\email{hw5x@virginia.edu}
\renewcommand{\shortauthors}{Li, Wu, and Wang}

\begin{abstract}
Collaborative bandit learning, i.e., bandit algorithms that utilize collaborative filtering techniques to improve sample efficiency in online interactive recommendation, has attracted much research attention as it enjoys the best of both worlds.
However, all existing collaborative bandit learning solutions impose a stationary assumption about the environment, i.e., both user preferences and the dependency among users are assumed static over time. Unfortunately, this assumption hardly holds in practice due to users' ever-changing interests and dependence relations, which inevitably costs a recommender system sub-optimal performance in practice.

In this work, we develop a collaborative dynamic bandit solution to handle a changing environment for recommendation. We explicitly model the underlying changes in both user preferences and their dependency relation as a stochastic process. Individual user's preference is modeled by a mixture of globally shared contextual bandit models with a Dirichlet Process prior. Collaboration among users is thus achieved via Bayesian inference over the global bandit models.
Model selection and arm selection for each user are done via Thompson sampling to balance exploitation and exploration. 
Our solution is proved to maintain a standard $\tilde O(\sqrt{T})$ sublinear regret even in such a challenging environment.  
And extensive empirical evaluations on both synthetic and real-world datasets further confirmed the necessity of modeling a changing environment and our algorithm's practical advantages against several state-of-the-art online learning solutions.
\end{abstract}




\maketitle

\section{INTRODUCTION}
Personalized recommendation is an essential component in most modern information service systems, as it helps alleviate information overload by tailoring the delivered content at a per-user basis \cite{resnick1997recommender,breese1998empirical}.
However, the content universe for most web services is usually large and undergoes frequent changes, which renders traditional methods, like collaborative filtering \cite{sarwar2001item,koren2009matrix} inappropriate, due to their offline training and online testing paradigm. 
Under this situation, the system needs to adaptively balance between the need of focusing on items that raise users' interest and the need of exploring new items for improving users' satisfaction in a long run. This exploration-exploitation dilemma is commonly formulated as a Multi-armed Bandit (MAB) problem \cite{UCB1}, and classical algorithms like upper confidence bound \cite{UCB1,LinUCB} and Thompson sampling \cite{pmlr-v54-abeille17a,Agrawal:2013:TSC:3042817.3043073} have been proved to be optimal in striking a balance between these two conflicting needs.
Therefore, bandit algorithms have become a reference solution to address this challenge. In particular, contextual bandit \cite{LinUCB}, an extension of MAB that incorporates contextual information, has been widely adopted in practice.

Moreover, as correlation between user preferences is common in many applications and contextual bandit cannot directly utilize it, various follow-up works seek to combine bandit algorithms with collaborative filtering in order to further improve sample efficiency via collaboration among user. For example, \cite{gentile2014online,li2019improved} performed online clustering of users in a bandit learning setting for collaborative filtering, and \cite{Li:2016:CFB:2911451.2911548,pmlr-v70-gentile17a} further considered context/arm-dependent clustering of users. In \cite{FactorUCB,kawale2015efficient} online matrix factorization is studied with bandit feedback. When social relation among users is available, such as social networks, the inferred user dependency is introduced as structured regularization for user-specific bandit model learning \cite{wu2016contextual,Gang,yang2020laplacian}.

We should note that all the existing collaborative bandit learning solutions impose a stationary assumption about the environment: they assume that both the user preferences and the dependency between users are static over time, which is a fundamental assumption in multi-armed bandit algorithms \cite{UCB1,Improved_Algorithm}. This unfortunately is often violated in real-world situations where users' preferences may change dramatically over time due to various internal or external factors \cite{wu2018dLinUCB,Hariri:2015:AUP:2832747.2832852}, which in turn lead to shifts in user dependencies \cite{tantipathananandh2007framework}. 
In some situations, non-stationarity may be alleviated to some extent by including contextual features describing external factors like season, topic and location, though it is usually difficult, if not impossible, to define such features ahead of time. However, this does not work when the non-stationarity is caused by internal factors of the users, which makes it necessary to design bandit algorithms that can adapt to such change in user preference. There have been a number of solutions proposed to address this challenge for MAB and contextual bandit problems \cite{garivier08_NonStationary, luo2018efficient, wu2018dLinUCB}, e.g., by detecting the change in user preference and then restarting the algorithm accordingly. However, to the best of our knowledge, none of the existing works has considered the more challenging problem of collaborative bandit learning in a changing environment, where both user preferences and their dependency relation become dynamic, giving rise to new challenges in arm selection, user clustering, and change detection.

In this work, we propose a new algorithm to address the aforementioned challenges in this new problem setting, i.e., collaborative bandit learning in a changing environment. Motivated by the social psychology theories about social norms \cite{doi:10.1177/001872675400700202} that humans tend to form groups with others of similar minds and ability, we explicitly model the underlying changes in \emph{both} user preferences and their dependency relation with a non-parametric stochastic process. Our solution does not assume an explicit network of users. Instead, we assume users share preference models in accordance of their underlying interest and dependency with others; and they switch models when their interest or received influence changes. To enable online learning of user preferences, we model the shared preference models with contextual linear bandits. Collaboration among users is thus achieved via Bayesian inference over the globally shared models. Model selection and arm selection in each user are performed via Thompson sampling \cite{pmlr-v54-abeille17a,Agrawal:2013:TSC:3042817.3043073} to balance exploitation and exploration. 
Our solution maintains a standard $\tilde O(\sqrt{T})$ sublinear regret even in such a challenging environment. Extensive empirical evaluations on both synthetic and real-world datasets for content recommendation confirmed the necessity of modeling a changing environment and our algorithm's practical advantages against several state-of-the-art online collaborative learning solutions.

\section{RELATED WORK}
Collaborative recommendation, including both traditional offline learning solutions such as collaborative filtering \cite{sarwar2001item,koren2009matrix}, and interactive online learning solutions, such as collaborative bandit learning \cite{Gang,wu2016contextual,FactorUCB,gentile2014online}, has shown great promise in personalized recommendation tasks. In particular, collaborative bandit learning, due to its ability of adapting to real-time user feedback, has received increasing attention in both industry and academia. 
Among them, there are several representative classes of solutions in modeling user dependency for collaborative recommendation. In the first type of solutions, when users' social relations are known (e.g., social network), the inferred dependency among users is encoded as a regularization for user-specific bandit model learning \cite{wu2016contextual,Gang,yang2020laplacian}. 
In the second type of solutions, where explicit user network is not assumed, the bandit parameters are estimated together with the dependency relation among users \cite{gentile2014online,pmlr-v70-gentile17a,li2019improved}. Typically, they cluster the user-specific bandit models via the learned model parameters during online updating. 
The third type of solutions appeal to latent factor models to capture the correlation between users and items in a lower dimensional space and estimate the latent factors with bandit feedback \cite{FactorUCB,kawale2015efficient}. We should note almost all existing collaborative learning solutions impose a stationary assumption about the environment, in which both user preferences and dependency are assumed to be static.

Non-stationarity appears in many real-world recommendation applications \cite{Moore2013TasteOT,Radinsky:2012:MPB:2187836.2187918}, and has shown to cost stationary recommendation algorithms sub-optimal performance \cite{wu2018dLinUCB}. In standard bandit learning settings, a number of solutions have been proposed to deal with non-stationarity for multi-armed bandit \cite{garivier08_NonStationary,auer2019adaptively},
contextual multi-armed bandit \cite{luo2018efficient, chen2019new}, and contextual linear bandit \cite{wu2018dLinUCB,cheung2019learning,russac2019weighted,zhao2020simple}. 
The main focus of these solutions is to eliminate the distortion from out-dated observations, which follow a different reward distribution than that of the current environment. To achieve this goal, common strategies include exponentially decaying the effect of past observations \cite{russac2019weighted}, discard past observations outside of a sliding window \cite{garivier08_NonStationary,cheung2019learning}, or adopt a change detector to actively detect the change point \cite{wu2018dLinUCB,yu2009piecewise} and then re-initialize the model.

However, these aforementioned solutions are not appropriate for collaborative recommendation in a non-stationary environment. First, none of the existing non-stationary bandit learning solutions model the possible dependency among users. This costs them the opportunity of leveraging the dependency among users to improve model estimation. Second, in online collaborative learning, not only individual users' preferences, but also the dependency among them, are subjected to unknown changes. Both factors have to be modeled for effective change detection and personalized recommendation.  In addition, these solutions are not sample efficient in the sense that they simply discard outdated models and observations, without reusing or sharing them with other users to improve model estimation at current time.



\section{Methodology}
In this section, we first introduce how to perform personalized interactive recommendation with contextual bandits in a stationary environment, which is the building block of our proposed collaborative dynamic bandit solution. Then we describe our non-parametric stochastic process model for modeling the dynamics in user preferences and dependency in a non-stationary environment. Finally, we provide the details about the proposed collaborative dynamic bandit algorithm and the corresponding theoretical regret analysis.     

\subsection{Contextual bandit for interactive recommendation} \label{subsec:contextual_bandit}
For online interactive recommendation, the system has to sequentially choose among a set of candidate items based on users' immediate feedback, such as click, ratings or dwell time \cite{LinUCB,wu2017returning}, in order to maximize the accumulated positive feedback in a finite period of time. This can be formulated as a contextual bandit problem \cite{LinUCB,Agrawal:2013:TSC:3042817.3043073},
where each candidate arm is associated with a $d$-dimensional feature vector $\bx$ referred to as the context (assume $\lVert \bx \rVert_2 \leq 1$ without loss of generality). Denote the candidate pool as $\cA_{t}=\{\bx_{t,1}, \bx_{t,2}, \dots, \bx_{t,|\cA_{t}|}\}$, which can be time-varying. The corresponding reward $r_t$ is governed by the context vector $\bx_t$ of the selected arm and an underlying \emph{fixed} but unknown bandit parameter $\btheta $ (assume $\lVert {\btheta} \rVert_2 \leq 1$).
In practice, a recommender system maintains one contextual bandit model $\btheta_u $ for each user $u$ for personalization \cite{LinUCB,Gang,wu2016contextual}.

Thompson Sampling (TS) \cite{pmlr-v54-abeille17a,Agrawal:2013:TSC:3042817.3043073} is a classic and popular bandit solution, which has been widely adopted in many real-world problems due to its flexibility and encouraging empirical performance. In TS, one needs to specify the prior distribution of the unknown bandit parameter $P(\btheta_{u})$ and the likelihood function of the reward $P(r_{i}|\bx_i, \btheta_u)$. Then with the set of observations $\{(\bx_{i},r_{i})\}_{i=1}^{t}$ collected so far, the posterior of $\btheta_{u}$ is obtained by $P(\btheta_{u}|\{(\bx_{i},r_{i})\}_{i=1}^{t}) \propto \prod^t_{i=1}P(r_{i}|\bx_{i},\btheta_{u}) P(\btheta_{u})$. 
With a linear reward assumption $P(r_{i}|\bx_i, \btheta_u)=\cN(r_{i}|\bx_{i}^{\top}\btheta_{u},\sigma^{2})$ and a conjugate prior $P(\btheta_{u})=\cN(\btheta_{u}|\mu_{0},\Sigma_{0})$, the posterior can be analytically computed as $P(\btheta_{u}|\{(\bx_{i},r_{i})\}_{i=1}^{t})=\cN(\btheta_{u}|\mu_{t},\Sigma_{t})$, where $\mu_{t}$ and $\Sigma_{t}$ denote the mean and covariance matrix respectively.
In each round $t$, TS samples the bandit parameter $\tilde \btheta_{u,t}$ from the posterior distribution, i.e., $\tilde \btheta_{u,t} \sim \cN(\btheta_{u}|\mu_{t-1},\Sigma_{t-1}) $, and then selects the arm with the highest reward under the sampled bandit parameter $\bx_t =  \argmax_{\bx \in\cA_t} \bx^{\top} \tilde \btheta_{u,t}$. In this work, we will restrict our attention to this linear reward setting. 


\subsection{Non-parametric modeling of an abruptly changing environment}
\label{sec_environment}

In this work, we consider a typical but non-trivial non-stationary environment, 
an abruptly changing environment \cite{garivier08_NonStationary,Hariri:2015:AUP:2832747.2832852, hartland:hal-00113668}, for each user in a collection of $N$ users, denoted as $\cU$. In this environment, the ground-truth bandit model $\btheta_{u,t}$ for a particular user changes arbitrarily at unknown time points in an \textit{asynchronous} manner, but remains constant between any two consecutive change points in this user. For example, in user $u$, we could have the following reward generation sequence,
\begin{equation*}
{\small
\!\!\!\!\!\!\underbrace{r_{0}, r_{1}, \!\cdots\!, r_{c_{u,1}-1}}_{ \text{governed by} ~  \btheta_{u, c_{u,0}}}, \underbrace{r_{c_{u,1}}, r_{c_{u,1}+1}, \!\cdots\!, r_{c_{u,2}-1}}_{  \text{governed by} ~ \btheta _{u, c_{u,1}}} , \!\cdots,\!\underbrace{r_{c_{u, \Gamma^u_T}}, r_{c_{u, \Gamma^{u}_{T}}+1},\!\cdots\!, r_{T}}_{ \text{governed by} ~\btheta _{u, c_{u,\Gamma^u_T}}}
\normalsize
}
\end{equation*}
where $c_{u,i}$ denotes the time step for the $i$-th change point of user $u$ (note that $c_{u,0}=0,\forall u\in \cU$). We should note that although the notations look verbose, the subscript $u$ on the change points is necessary because the changes in different users are not necessarily synchronized. $\btheta _{u,c_{u,i}}$ is the ground-truth bandit parameter for user $u$ between his/her $i$-th and the $(i+1)$-th change point. The change points $\cC_{u,T}=\{c_{u,i}\}_{i \in [0,\Gamma_{T}^{u}]}$ of the underlying reward distribution for user $u \in \cU$ up to time $T$ and the corresponding bandit parameters $\Theta_{u,T}=\{ \btheta _{u, c}\}_{c \in \cC_{u,T}}$ are unknown to the learner. $\Gamma^u_T $ denotes the number of change points for user $u$ up to time $T$, which is also unknown. To reflect the nature of a collaborative learning environment, we further assume the bandit parameters $\Theta_{u,T}$ in each user overlap across the $N$ users. Therefore, at a particular moment, users who share the same bandit parameters form clusters; and of course, this clustering structure is unknown to the learner as well. Due to the asynchronous changes of bandit parameters among users, the clustering of users is also evolving over time. 


In such a non-stationary environment, existing contextual bandit solutions become incompetent, as the accumulated observations across different stationary periods damage their parameter estimation quality. 
Existing solutions \cite{wu2018dLinUCB,Hariri:2015:AUP:2832747.2832852} concerning such an environment detect the changes in each user \emph{independently} and re-build their parameter estimation from scratch after each detected change point. This unfortunately ignores the fact that users are related to each other in such a changing environment, e.g., the dynamically formed user clusters.
In the rest of this section, we describe how we explicitly model the change in users as a stochastic process, which brings in the possibility of dynamic collaborative learning.

Motivated by the social psychology theories about social norms \cite{doi:10.1177/001872675400700202}, in this work instead of considering the preferences of each user as fixed but unknown, we treat them as stochastic by assuming each user's model parameter $\btheta_{u,c}$ is drawn from a Dirichlet Process (DP) \cite{antoniak1974,ferguson1973}. Specifically, a Dirichlet Process, DP($\alpha_{0}$, $G_0$) with a base distribution $G_0$
and a scaling parameter $\alpha_{0}$, is a distribution over distributions. An
important property of DP is that samples from it often share some common values, and therefore naturally form clusters. The number
of unique draws, i.e., the number of clusters, varies with respect to
the data and thus is random, instead of being pre-specified. 
This process can be formally described as follows,
\begin{align} \label{eq:DP}
     G &\sim  \text{DP}(\alpha_{0}, G_0) \\ \nonumber
    \btheta_{u, c_{u,i}} | G & \sim  G, \forall u\in \cU, c_{u,i} \in \cC_{u} \\ \nonumber
     r_{t} | \btheta_{u, c_{u,i}} , \bx_t & \sim \cN(r_{t}| \bx_t^\top\btheta_{u, c_{u,i}},\sigma^2), \forall t \in [c_{u,i},c_{u,i+1}-1]  \nonumber
\end{align}
where the hyper-parameter $\alpha_{0}$ controls the concentration of unique draws from the DP prior, the base distribution $G_0$ specifies the prior distribution of the bandit parameters in each individual model, and $G$ represents the mixing distribution of the sampled results of $\btheta_{u, c}$. To enable efficient posterior inference, conjugate priors are expected in $G_0$. Due to our linear reward assumption, we impose a zero-mean isotropic Gaussian prior governed by a single precision parameter $\lambda$ on $\btheta_{u,c} $ as $G_0 = N(\mathbf{0}, \lambda^{-1} \bI)$. 
With the DP prior defined above, when a new user arrives or an existing user changes his/her preference at time $t$, the distribution of this user's new bandit parameter $\btheta_{u,t}$ conditioned on all existing bandit parameters $\Theta_{t-1}=\{ \btheta _{u, c}\}_{u \in \cU, c \in \cC_{u,t-1}}$ can be analytically derived by integrating out $G$ in Eq \eqref{eq:DP}:
\begin{align} \label{eq:Pólya_urn}
    P(\btheta_{u,t}|\Theta_{t-1}, \alpha_{0}, G_0) = \frac{\alpha_{0} G_0}{|\Theta_{t-1}|+\alpha_{0}} + \frac{\sum_{\btheta \in \Theta_{t-1}} \delta_{\btheta_{u,t}}(\btheta)}{|\Theta_{t-1}|+\alpha_{0}}
\end{align}
where $\delta_{\btheta_{u,t}}(\cdot)$ is a delta function concentrated at $\btheta_{u,t}$. This conditional distribution well captures the idea of social psychology theories about social norms \cite{doi:10.1177/001872675400700202}: when a user's preference changes or a new user comes, the prior distribution over the new model that he/she tends to choose is proportional to the popularity of existing models in overall user population at the moment.

To facilitate our discussion about this clustering property, we denote the set of unique draws in $\Theta_{t-1}$ as $\{\phi_{z}\}_{z=1}^{K_{t-1}}$, where $K_{t-1}$ is the total number of unique draws from DP so far. Then we introduce an indicator variable $z_{u,t}$ such that $\btheta_{u,t}=\phi_{z_{u,t}}$, i.e., $z_{u,t}$ is the model index in this globally shared unique bandit parameter set. Denote $\cZ_{t}=\{ z_{u,c}\}_{u \in \cU, c \in \cC_{u,t}}$, and an equivalent form of Eq \eqref{eq:Pólya_urn} is:
\begin{equation} \label{eq:CRP}
  P(z_{u,t}=k|\alpha_{0}, \cZ_{t-1}\}) \propto
    \begin{cases}
      {n_{k,t-1}} & \text{if $k\in[K_{t-1}]$}\\
      {\alpha_{0}} & \text{if $k = K_{t-1}+1$}\\
    \end{cases}   
\end{equation}
where $n_{k,t-1}=\sum_{z \in \cZ_{t-1}} \bm{1}\{z=k\}$ is the number of times elements in $\Theta_{t-1}$ takes value $\phi_{k}$.

As a result, the imposed DP prior encourages users to form shared groups at any particular moment of time, which makes online collaborative learning feasible.
We should emphasize that our collaborative bandit solution does not require any knowledge about $K_T$ or $\{\phi_{z}\}_{z=1}^{K_T}$, but adaptively learns them via Bayesian inference with the observations obtained during its interaction with users. 

\subsection{Collaborative Dynamic Bandit}

\begin{figure}[t] 
\centering
\includegraphics[width=0.92\linewidth]{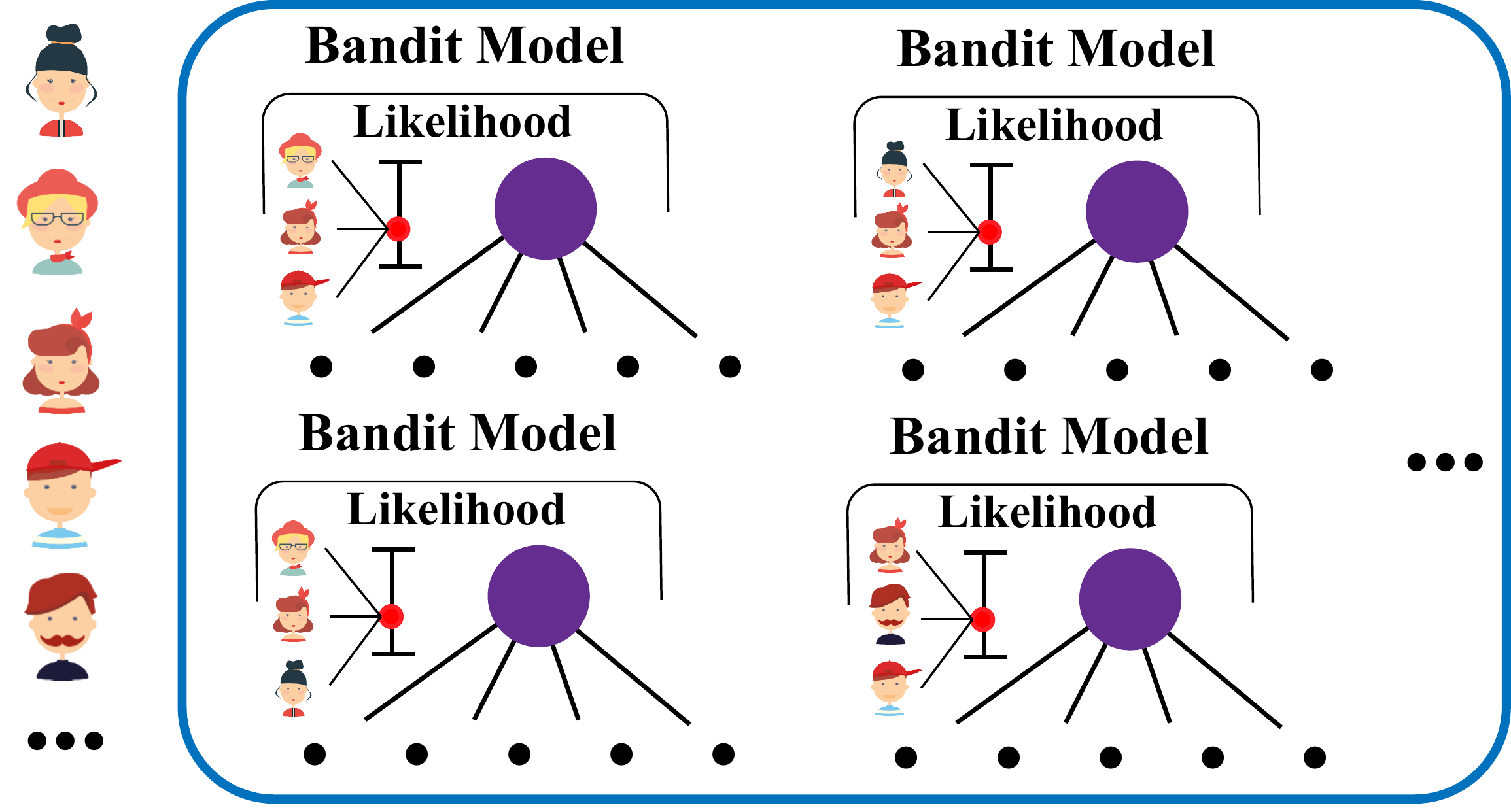}
\vspace{-3mm}
\caption{Illustration of CoDBand. An adaptively maintained pool of contextual bandit models is shared among all the users with respect to the underlying clustering structure of them. Bandit models are assigned to users based on fitness with user history data.} \label{fig:illustrative_fig}
\vspace{-4mm}
\end{figure}

In the non-stationary environment specified above, to make personalized recommendations in real-time, several challenges have to be addressed: 1) as the changes in a user are unknown to the learner, how to detect the potential changes in each user's bandit parameters; 2) how to estimate the globally shared bandit parameters with the observations obtained from different users. 

As our solution, an adaptively maintained pool of contextual bandit models is shared among all the users (as shown in Figure \ref{fig:illustrative_fig}).
To address the challenges above, a change point detector is used to detect the changes in each user's bandit parameter, and a collapsed Gibbs sampler is used to select a suitable bandit model to serve the user. This sampling procedure selects a global bandit model for a user by taking into consideration both how well the bandit model fits the user's recent historical data as well as the model's popularity among all the users. This captures the intuition that when there is limited knowledge about a user (e.g., cold start), it is better to explore whether the well-established popular models fit the user, compared with directly starting from scratch (as in \cite{wu2018dLinUCB}). Global bandit models are created, updated or removed from the pool in an adpative manner as the algorithm interacts with the users.
We name the resulting bandit algorithm as Collaborative Dynamic Bandit, or CoDBand in short, and illustrate the details of it in Algorithm \ref{alg:CoDBand}. 

\begin{algorithm}[h]
\SetAlgoLined
\LinesNumbered
\SetKwInOut{Input}{Input}\SetKwInOut{Initialize}{Initialize}\SetKwInOut{Output}{Output}%
\caption{Collaborative Dynamic Bandit (CoDBand)} \label{alg:CoDBand}
\Input{$\sigma$, $a$, $b$, $\lambda$, $\delta_{1}$, $\delta_{2}$, $\tau$}
\Initialize
{\begin{minipage}[t]{6cm}%
 \strut
 Construct user set $\cU$ and initialize $\cU = \emptyset$.
 
 Construct global bandit model set $\cG$ and initialize $\cG = \emptyset$. Sample $\alpha_{0} \sim \Gamma(a,b)$.
 \strut
\end{minipage}%
}
\For{$t=1$ to $T$}{
    Observe current user $u_t$, and candidate arm pool $\cA_t$\;
    \If{$u_t \notin \cU $}{
        $\cU = \cU \cup u_t$\;
        Initialize an observation set for $u_t$: $\cD_{t-1}^{u_t} = \emptyset$;
    }
    \textbf{ARM SELECTION}\;
    \If{$\cD_{t-1}^{u_t} = \emptyset$}{
        Sample a model index $\tilde z_{u_{t}}$ for user $u_{t}$ using Eq \eqref{eq:CRP} \;

        \If{$\tilde z_{u_{t}}  = |\cG|+1 $}{
            Initialize a new global model $\cM_{\tilde z_{u_{t}}}$: $n_{\tilde z_{u_{t}}}=0$, $\Sigma^{-1}_{\tilde z_{u_{t}}}=\lambda I \in \bbR^{d \times d}$, $\bb_{\tilde z_{u_{t}}}=\textbf{0} \in \bbR^{d}$, $\mu_{\tilde z_{u_{t}}}=\Sigma_{\tilde z_{u_{t}}}\bb_{\tilde z_{u_{t}}}$\;
            Add it to the global model set $\cG = \cG \cup \cM_{\tilde z_{u_{t}}}$\;
        }
        $n_{\tilde z_{u_{t}}}=n_{\tilde z_{u_{t}}}+1$ \;
    }
    Sample $\tilde{\theta}_{t} \sim \cN(\mu_{\tilde z_{u_{t}}}, \Sigma_{\tilde z_{u_{t}}})$\;
	Select $x_{t}=\argmax_{x\in\cA_t} \bx^{\top} \tilde{\theta}_{t}$, and observe reward $r_{t}$\;
	\textbf{MODEL UPDATE}\;
	Compute $e_{u_{t},t}$ according to Eq \eqref{eq:badness}, and update $\hat{e}_{u_{t},t}$\;
	Update the observation set: $\cD_{t}^{u_t} = \cD^{u_{t}}_{t-1} \cup \{(\bx_{t}, r_{t})\}$\;
	Update global model $\cM_{\tilde z_{u_{t}}}$: $\Sigma^{-1}_{\tilde z_{u_{t}}}=\Sigma^{-1}_{\tilde z_{u_{t}}}+\frac{1}{\sigma^{2}}\bx_{t}\bx_{t}^{\top}$, $\bb_{\tilde z_{u_{t}}}=\bb_{\tilde z_{u_{t}}}+\frac{1}{\sigma^{2}}\bx_{t} r_{t}$, $\mu_{\tilde z_{u_{t}}}=\Sigma_{\tilde z_{u_{t}}}\bb_{\tilde z_{u_{t}}}$\;
	$\tilde z_{u_{t}}, \cG=\text{Collapsed Gibbs Sampler}(\tilde z_{u_{t}}, \cD^{u_t}_{t}, \cG)$ \;
	$\alpha_{0}=\text{Update Parameter}(\alpha_{0}, |\cG|, a, b, \sum_{k=1}^{|\cG|}n_{k})$ \cite{escobar1995bayesian}\;
	\textbf{CHANGE DETECTION}\;
	\If{$\hat{e}_{u_{t},t} > \delta_{1} + \sqrt{\frac{\log{1/\delta_{2}}}{\tau}}$}{
	Set $\cD_{t}^{u_t} = \emptyset$, $\hat{e}_{u_{t},t}=0$ \;
	}
}
\end{algorithm}

Before presenting the detailed description of the two core components of CoDBand, i.e., change detection and collapsed Gibbs sampling, we first introduce how observations are managed in it:
\begin{itemize}
    \item CoDBand maintains a set $\cD_{t}^{u}$ for each user $u \in \cU$ that is updated by each new observation from $u$ (line 19 in Algorithm \ref{alg:CoDBand}), and is reset to $\cD_{t}^{u}=\emptyset$ when a change point in $u$ is detected (line 25-26 in Algorithm \ref{alg:CoDBand}). As a result, $\cD_{t}^{u}$ reflects the target user $u$'s recent preferences, as it only contains observations in the current stationary period of $u$ with a high probability.
    \item CoDBand also maintains a pool of globally shared bandit models denoted as $\cG_{t}$, and each bandit model $\cM_{k,t} \in \cG_{t}$ maintains a posterior distribution $\cN(\mu_{k,t}, \Sigma_{k,t})$ of the unknown bandit parameter and a counter $n_{k,t}$ recording the number of times $\cM_{k,t}$ is assigned to a user (line 14 in Algorithm \ref{alg:CoDBand}). It is obvious from the context that $\cG_{t}$, $\cM_{k,t}$, $\mu_{k,t}$, $\Sigma_{k,t}$ and $n_{k,t}$ are all updated over time, so the subscript $t$ is omitted for simplicity in the following discussions.
\end{itemize}

Intuitively, each bandit model $\cM_{k} \in \cG$ represents a typical type of user behaviors that are learned from the system's interaction history with all users. The set $\cD_{t}^{u}$ serves as an anchor to decide which bandit model $\cM_{k}$ best fits user $u$'s recent preferences. In the rest of this section, we will introduce details about how we perform change detection to maintain $\cD_{t}^{u}$, and how we use collapsed Gibbs sampling to update and select $\cM_{k}$ in individual users.

\subsubsection{Change Detection} \label{subsubsec:cd}
Since we assume change points are arbitrary and unknown to the learner, the change point detector from \cite{wu2018dLinUCB} can be adopted to detect the changes in a user's bandit parameter.
This is done by constructing the test variable
\begin{equation}\label{eq:badness}
e_{u_{t},t}=\textbf{1}\bigl\{|\hat{r}_{t}-r_{t}|>\text{CB}_{u_{t},t-1}(\bx_{t})+\epsilon \bigr\}.    
\end{equation}
$e_{u_{t},t}$ indicates whether the received reward $r_{t}$ deviates too much from the estimated reward $\hat{r}_{t}=\bx^{\top}\hat{\theta}_{u_{t},t-1}$, where $\hat{\theta}_{u_{t},t-1}=\big(\lambda I +\sum_{(\bx_{i},r_{i}) \in \cD_{t-1}^{u_{t}}}\bx_{i}\bx_{i}^{\top}\big)^{-1} \big(\sum_{(\bx_{i},r_{i}) \in \cD_{t-1}^{u_{t}}}r_{i}\bx_{i}\big)$ is the Ridge regression estimator using observations in $\cD_{t-1}^{u_{t}}$. $\text{CB}_{u_{t},t-1}(\bx)$ denotes the high probability confidence bound from \cite{Improved_Algorithm}, which is defined as $\text{CB}_{u_{t},t-1}(\bx)=\alpha_{u_{t},t-1}\sqrt{\bx^{\top} \Big(\lambda I +\sum_{(\bx_{i},r_{i}) \in \cD_{t-1}^{u_{t}}}\bx_{i}\bx_{i}^{\top}\Big)^{-1}\bx}$,
where $\alpha_{u_{t},t-1}=\sigma\sqrt{d\log{\Big(1+\frac{|\cD_{t-1}^{u_{t}}|}{d\lambda}\Big)} + 2\log{\frac{1}{\delta_{1}}}}+\sqrt{\lambda}$. And in Eq \eqref{eq:badness}, $\epsilon=\sqrt{2}\sigma \text{erf}^{-1}(\delta_{1}-1)$ represents the high probability bound of Gaussian noise in the received feedback and $\text{erf}^{-1}(\cdot)$ is the inverse of Gaussian error function. 

When the reward distribution remains stationary (e.g., observations in $\cD_{t-1}^{u_{t}}$ and $(\bx_{t},r_{t})$ are homogeneous), with a probability at least $1-\delta_{1}$, the test variable $e_{u_{t},t}=0$ \cite{wu2018dLinUCB}. To account for the noise in one individual observation, an empirical mean of $e_{u_{t},t}$ over the $\tau$ most recent interactions with user $u_{t}$ is maintained, which is denoted as $\hat{e}_{u_{t},t}=\frac{1}{\min(|\cD_{u_{t},t-1}|,\tau)}\sum_{i}e_{u_{t},i}$ (line 19 in Algorithm \ref{alg:CoDBand}).
When $\hat{e}_{u_{t},t} > \delta_{1} + \sqrt{\frac{\ln{1/\delta_{2}}}{2\tau}}$ (obtained by Hoeffding inequality), a change is said to be detected in user $u_{t}$'s bandit parameter and the value of $\hat{e}_{u_{t},t}$ is reset (line 25-26 in Algorithm \ref{alg:CoDBand}).

\subsubsection{Collapsed Gibbs Sampling} \label{subsubsec:gibbssampling}
As mentioned earlier, a collapsed Gibbs sampler is used to select the bandit model $\cM_{k}\in \cG$ for user $u$, by sampling a model index $\tilde z_u$ from its posterior distribution conditioned on $\cD_{t}^{u}$, as illustrated in Algorithm \ref{alg:GibbsSampler}. The conditional posterior of $\tilde z_{u}$ consists of two parts: the conditional prior distribution of $\tilde z_{u}$ in Eq \eqref{eq:CRP}, e.g., popularity of the bandit model among all users, and the marginalized likelihood on $\cD_{t}^{u}$, e.g., fitness with the user's recent history. With the conjugate Gaussian prior we introduced in Eq \eqref{eq:DP}, the marginalized likelihood $P(r_{i}|\bx_{i},\tilde z_{u}=k, \cG) = \int \cN\big(r_i| \bx_i^\mt \phi, \sigma^2\big)  \cN\big(\phi | \mu_{k}, \Sigma_{k}^{-1}\big) d\phi = \cN\big(r_i| \bx_i^{\mt} \mu_{k}, \sigma^2 + \bx_i^\mt \Sigma_{k}^{-1} \bx_i\big)$ can be analytically computed. Therefore, the conditional posterior distribution of $\tilde z_{u}$ can be computed as,
\small
\begin{equation} \label{eq:posterior_predictive}
\begin{split}
& P(\tilde z_{u} = k|\alpha_{0}, \{n_{k}\}_{k=1}^{K}, \cD^{u}, \cG) \\
& \propto P(\tilde z_{u}=k|\alpha_{0}, \{n_{k}\}_{k=1}^{K}) P(\cD^{u}|\tilde z_{u}=k, \cG) \\
& \propto
    \begin{cases}
    {n_{k}} \prod_{(\bx_i, r_i) \in \cD^u} \cN\big(r_i| \bx_i^{\mt} \mu_{k}, \sigma^2 + \bx_i^\mt \Sigma_{k}^{-1} \bx_i\big) & \text{if $k\in[K]$}\\
      \alpha_{0} \prod_{(\bx_i, r_i) \in \cD^u} \cN\big(r_i| 0, \sigma^2 + \lambda^{-1} \bx_i^\mt \bx_i\big) & \text{if $k = K+1$}
    \end{cases}  
\end{split}
\end{equation}
\normalsize

In addition, we add a Gamma prior on the concentration parameter $\alpha_{0} \sim \Gamma(a,b)$ and update it with Gibbs sampling as well. The sampling procedure for $\alpha_{0}$ follows Section 6 of \cite{escobar1995bayesian}. Therefore, instead of manually tuning $\alpha_{0}$, we can estimate it during the interactions with users (line 23 in Algorithm \ref{alg:CoDBand}).

\begin{algorithm}[t]
\SetAlgoLined
\LinesNumbered
\SetKwInOut{Input}{Input}\SetKwInOut{Initialize}{Initialize}\SetKwInOut{Output}{Output}%
\caption{Collapsed Gibbs Sampler} \label{alg:GibbsSampler}
\Input{model index $\tilde z_{u}$, observation set $\cD^{u}$, global bandit model set $\cG$}
\Output{new model index $\tilde{z}_{u}^{\prime}$, updated global bandit model set $\cG$}

Remove $\cD^{u}$ from global model $\cM_{\tilde z_{u}}$: $n_{\tilde z_{u}} = n_{\tilde z_{u}} - 1$, $\Sigma^{-1}_{\tilde z_{u}}=\Sigma^{-1}_{\tilde z_{u}}-\frac{1}{\sigma^{2}}\sum_{(x_{i},r_{i}) \in \cD^{u}}\bx_{i}\bx_{i}^{\top}$, $\bb_{\tilde z_{u}}=\bb_{\tilde z_{u}}- \frac{1}{\sigma^{2}}\sum_{(x_{i},r_{i}) \in \cD^{u}}\bx_{i} r_{i}$, $\mu_{\tilde z_{u}}=\Sigma_{\tilde z_{u}}\bb_{\tilde z_{u}}$\;

\If{$n_{\tilde z_{u}} = 0$}{
    Remove $\cM_{\tilde z_{u}}$: $\cG = \cG \setminus \cM_{\tilde z_{u}}$
}

Sample new model index $\tilde{z}_{u}^{\prime}$ for $\cD^{u}$ according to Eq \eqref{eq:posterior_predictive}\;

Update global model $\cM_{\tilde z_{u}^{\prime}}$ with $\cD^{u}$: $n_{\tilde z_{u}^{\prime}} = n_{\tilde z_{u}^{\prime}} + 1$, $\Sigma^{-1}_{\tilde z_{u}^{\prime}}=\Sigma^{-1}_{\tilde z_{u}^{\prime}}+\frac{1}{\sigma^{2}}\sum_{(x_{i},r_{i}) \in \cD^{u}}\bx_{i}\bx_{i}^{\top}$, $\bb_{\tilde z_{u}^{\prime}}=\bb_{\tilde z_{u}^{\prime}}+\frac{1}{\sigma^{2}}\sum_{(x_{i},r_{i}) \in \cD^{u}}\bx_{i} r_{i}$, $\mu_{\tilde z_{u}^{\prime}}=\Sigma_{\tilde z_{u}^{\prime}}\bb_{\tilde z_{u}^{\prime}}$\;

\end{algorithm}

In the model update stage of each iteration (line 22 in Algorithm \ref{alg:CoDBand}), the collapsed Gibbs sampler is executed to re-assign the model index for the user $u_{t}$ given this user's $\cD_{t}^{u_{t}}$, and the bandit models involved in this procedure will be updated accordingly (line 1 and 6 in Algorithm \ref{alg:GibbsSampler}). Intuitively, as we have more observations about the user, we can select a better suited global model for him/her with an increasing confidence. 

It is worth noting that when the target user is new or with newly detected changes, CoDBand tends to choose a currently popular model for him/her to start with (line 8-9 in Algorithm \ref{alg:CoDBand}), rather than to always create a new model, due to our underlying DP modeling assumption of user preferences. This choice is arguably preferred, especially when a large population of users are presented. As the sufficient statistics are maintained at the model-level, e.g., the globally shared models in $\cG$, rather than at the user-level, collaborative learning is achieved. When a user switches to an existing model, the system can take advantage of the already accumulated statistics to make more accurate recommendations for this user.

After a model is sampled for the user, arm selection is conducted using Thompson sampling (line 16-17 in Algorithm \ref{alg:CoDBand}). Compared with standard Thompson sampling \cite{Agrawal:2013:TSC:3042817.3043073,chapelle2011empirical}, we are introducing another layer of exploration in the model space. This is because CoDBand first samples a model index from the posterior over all possible bandit models and then samples a bandit parameter conditioning on the sampled model.

\section{Regret Analysis}

We analyze the accumulative Bayesian regret of CoDBand, which is defined as:
\begin{equation} \label{eq:bayesian_regret}
    \bbE[R_{T}] = \bbE\bigl[\sum_{t=1}^{T}r_{t}\bigr] = \bbE\bigl[\sum_{t=1}^{T}{\bx^{*}_{t}}^{\top} \theta_{u_{t},t}-\bx_{t}^{\top}\theta_{u_{t},t}\bigr],
\end{equation}
where the expectation is taken with respect to the prior distribution of the bandit parameter $\theta_{u_{t},t}$. $\bx_{t}^{*}$ is the best arm in hindsight and $\bx_{t}$ is the selected arm at time $t$.
To analyze Bayesian regret, we define the upper confidence bound function $U_{t}: [K_{t}] \times \mathcal{A}_{t} \rightarrow \mathbb{R}$ and the lower confidence function $L_{t}: [K_{t}] \times \mathcal{A}_{t} \rightarrow \mathbb{R}$ by
\begin{align*}
    & U_{t}(k,\bx) = f_{\hat{\theta}_{k,t-1}}(\bx) + \alpha_{k,t-1}||\bx||_{V_{k,t-1}^{-1}} \\
    & L_{t}(k,\bx) = f_{\hat{\theta}_{k,t-1}}(\bx) - \alpha_{k,t-1}||\bx||_{V_{k,t-1}^{-1}}
\end{align*}
where $V_{k,t}=\lambda I+\sum_{s \in \cI(k)}\bx_{i}\bx_{i}^{\top}$, and $\cI(k)$ denotes the set of time steps where the bandit parameter $\theta_{u_{t},t}$ takes value $\phi_{k}$.

Denote $\cH_{t}=\sigma(\bx_{1},r_{1},\dots,\bx_{t},r_{t})$ as the $\sigma$-algebra generated by the interaction sequence at time step $t$. Our regret analysis draws its key insight from \cite{russo2014learning} that for Thompson sampling method we have $\bbP(\tilde{z}_{t}|\cH_{t-1})=\bbP(z^{*}_{t}|\cH_{t-1})$ and $\bbP(\bx_{t}|\tilde{z}_{t}=k,\cH_{t-1})=\bbP(\bx_{t}^{*}|z^{*}_{t}=k,\cH_{t-1})$. Therefore, $\bbP[(\bx_{t}=x)\cap(\tilde{z}_{t}=k)|\cH_{t-1}]=\bbP[(\bx_{t}^{*}=x)\cap(z^{*}_{t}=k)|\cH_{t-1}]$. In addition, 
since $U_{t}(k,\bx)$ and $L_{t}(k,\bx)$ are deterministic functions, $\bbE[U_{t}(\tilde{z}_{t},\bx_{t})|\cH_{t-1}]=\bbE[U_{t}(z^{*}_{t},\bx_{t}^{*})|\cH_{t-1}]$ and $\bbE[L_{t}(\tilde{z}_{t},\bx_{t})|\cH_{t-1}]=\bbE[L_{t}(z^{*}_{t},\bx_{t}^{*})|\cH_{t-1}]$. Based on a similar decomposition as in \cite{russo2014learning,lattimore2020bandit}, we obtain the following result.

\begin{lemma} \label{lem:regret_decomposition}
The accumulated Bayesian regret defined in Eq \eqref{eq:bayesian_regret} can be decomposed into the following three terms:
\small
\begin{equation*}
\begin{split}
    \bbE[R_{T}] & \leq 2 \sum_{t=1}^{T}\bbP\{ \bigl[ \bx_{t}^{\top}\theta_{u_{t},t} < L_{t}(z^{*}_{t}, \bx_{t})\bigr] \cup \bigl[ {\bx^{*}_{t}}^{\top}\theta_{u_{t},t} > U_{t}(z^{*}_{t},\bx_{t}^{*})\bigr] \} \\
    & +\sum_{t=1}^{T}\bbE[U_{t}(z^{*}_{t}, \bx_{t})-L_{t}(z^{*}_{t}, \bx_{t})] \\
        & + \sum_{t=1}^{T}\bbE[\bigl[U_{t}(z^{*}_{t}, \bx_{t}^{*})-U_{t}(z^{*}_{t}, \bx_{t})\bigr]\cdot \mathbf{1}\bigl\{\tilde{z}_{t} \neq z^{*}_{t}\bigr\}]
\end{split}
\end{equation*}
\normalsize
\end{lemma}

It is worth noting that the first two terms can be found in the Bayesian regret for linear Thompson sampling (Section 6.2.1 in \cite{russo2014learning}) as well: the first term is related to the case when reward estimation error exceeds its high confidence bound, which is bounded by the constant $4$ based on Theorem 2 in \cite{Improved_Algorithm}; the second term corresponds to the rate of convergence of the confidence interval. and by rewriting the summation over each model, and then applying Theorem 3 in \cite{Improved_Algorithm}, it is bounded by $O\Big(\sigma d\sqrt{T}\log{T}(\sum_{k=1}^{K_{T}}\sqrt{p_{k}})\Big)$ where $p_{k}=\frac{|\cI(k)|}{T}$ for $k\in[K_{T}]$ denotes the portion of time steps that the bandit parameter takes value $\phi_{k}$.



The key difference between our regret analysis and that of linear Thompson sampling is the additional third term, which corresponds to the regret due to sampling a wrong model. This is unique to our problem because compared with linear Thompson sampling, CoDBand addresses exploration and exploitation not only on arm level, but also on model level. To bound this term, we further decompose it based on whether late detection has happened.
Denote $\cL_{t}$ as the late detection event at time $t$ that the change detector defined in Section \ref{subsubsec:cd} fails to detect the most recent change point so far, and the complement of $\cL_{t}$ is denoted as $\cL^{C}_{t}$.  Then we can further decompose the third term as:
\small
\begin{align*}
    & \sum_{t=1}^{T}\bbE[\bigl[U_{t}(z^{*}_{t}, \bx_{t}^{*})-U_{t}(z^{*}_{t}, \bx_{t})\bigr]\cdot \mathbf{1}\bigl\{\tilde{z}_{t} \neq z^{*}_{t}\bigr\}]  \leq C_{0}\sum_{t=1}^{T}\bbE[\mathbf{1}\bigl\{\tilde{z}_{t} \neq z^{*}_{t}\bigr\}] \\
    & = C_{0}\sum_{t=1}^{T}P(\tilde{z}_{t} \neq z^{*}_{t}|\cL_{t}^{C}) P(\cL_{t}^{C}) + C_{0}\sum_{t=1}^{T}P(\tilde{z}_{t} \neq z^{*}_{t}|\cL_{t}) P(\cL_{t})\\
    & \leq \underbrace{C_{0}\sum_{t=1}^{T} P(\cL_{t})}_{A_{1}} + \underbrace{C_{0}\sum_{t=1}^{T}P(\tilde{z}_{t} \neq z^{*}_{t}|\cL_{t}^{C})}_{A_{2}}
\end{align*}
\normalsize
where $C_{0}=2+\sigma \sqrt{\frac{2}{\lambda}\log{\frac{1}{\delta_{1}}}}$ is the constant upper bound of $U_{t}(z^{*}_{t}, \bx_{t}^{*})$ obtained by setting $t=0$.

The term $A_{1}$ represents the penalty in regret due to late detection; and the following lemma provides an upper bound of it.

\begin{lemma}
Let $S_{u,c}$ denote the length of stationary period after the $c$'th change point of user $u$. According to Lemma 3.4 in \cite{wu2018dLinUCB}, assume at least $\rho$ portion of arms in $\cA_{t},\forall t$ satisfy $|\bx^{\top}\theta_{u,c_{u,i}}-\bx^{\top}\theta_{u,c_{u,i+1}}| \geq \Delta$, and by setting $\delta_{1}\leq 1-\frac{1}{\rho}\big(1-\frac{\sqrt{\lambda}}{2\min(S_{u,c})\rho}(\Delta-2\sqrt{\lambda}-2\epsilon)\big)$ and $\tau \geq \frac{2\ln{\frac{2}{\delta_{2}}}}{(\rho(1-\delta_{1})-\delta_{1})^{2}}$, the probability of detection when change has happened is $p_{d} \geq 1-\delta_{2}$. This leads to the following upper bound of the term $A_{1}$:
\small
\begin{align*}
     A_{1} &=C_{0} \sum_{u \in \cU} \sum_{c \in \cC_{u,T}} \sum_{t \in S_{u,c}} P(\cL_{t}) \leq C_{0}\sum_{u \in \cU} \sum_{c \in \cC_{u,T}} \frac{1-\delta_{2}^{|S_{u,c}|}}{1-\delta_{2}} \\
    & \leq  \frac{C_{0}}{1-\delta_{2}}\sum_{u \in \cU_{T}} \Gamma_{T}^{u}
\end{align*}
\normalsize
\end{lemma}

The term $A_{2}$ corresponds to the penalty in regret caused by sampling a wrong model for arm selection when there is no late detection. It is related to the reward gap $\Delta$ between different bandit parameters as well as the model's confidence in the estimation. We bound it by the following lemma.
\begin{lemma} \label{lemma:wrong_model}
Adopting the same assumption as in \citep{gentile2014online,pmlr-v70-gentile17a}, at each time $t$, arm set $\cA_{t}$ is generated i.i.d. from a sub-Gaussian random vector $X \in \bbR^{d}$, such that $\mathbb{E}[XX^{\top}]$ is full-rank with minimum eigenvalue $\lambda'>0$; and the variance $\varsigma^{2}$ of the random vectors satisfies $\varsigma^{2} \leq \frac{{\lambda'}^{2}}{8\ln{4K}}$. Then the term $A_{2}$ can be upper bounded by:
\begin{align*}
    A_{2} = C_{0}\sum_{t=1}^{T}P(\tilde{z}_{t} \neq z^{*}_{t}|\cL_{t}^{C}) = O(K_{T} \bigl[(\sum_{u \in \cU} \Gamma^{u}_{T})+C_{1}\bigr])
\end{align*}
with probability at least $1-\delta^{'}$, where $C_{1}=\frac{\psi_{L}\sigma^{2}}{\Delta^{2}{\lambda^{'}}^{2}}\log{\frac{d}{\delta^{'}}}$ and $\psi_{L}$ is a constant that depends on $d,\sigma,\lambda$.
\end{lemma}

\begin{figure*}[t]
\centering
\begin{tabular}{c c c}
\includegraphics[width=5.7cm]{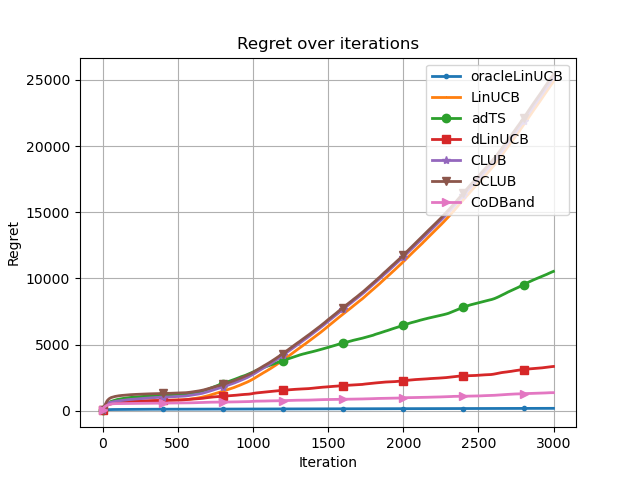} &
\includegraphics[width=5.7cm]{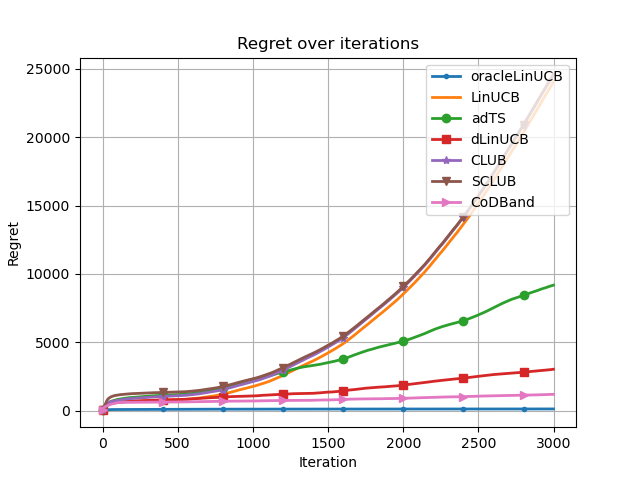} &
\includegraphics[width=5.7cm]{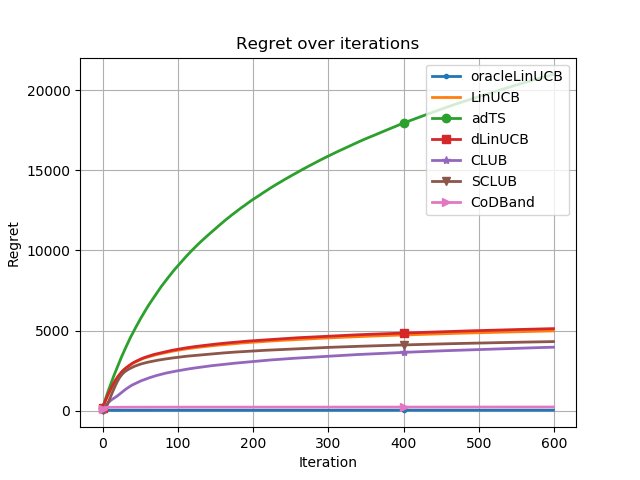} \\
\small (a) Simulation setting 1  & \small (b) Simulation setting 2  & \small (c) Simulation setting 3
\normalsize
\end{tabular}
\vspace{-2mm}
\caption{Performance comparison on synthetic datasets.} \label{fig:simulation}
\vspace{-1mm}
\end{figure*}

Combining all the components together we obtain the final regret upper bound $\bbE[R_{T}] = O\Big(\sigma d\sqrt{T}\log{T}(\sum_{k=1}^{K_{T}}\sqrt{p_{k}}) + K_{T}\sum_{u \in \cU} \Gamma_{T}^{u}\Big)$. CoDBand achieves a standard $\tilde{O}(\sqrt{T})$ regret bound with respect to time horizon $T$, and the added regret only depends on the underlying grouping structure among users $\sum_{k=1}^{K_{T}}\sqrt{p_{k}}$ and the total number of stationary periods among all users $\sum_{u \in \cU} \Gamma_{T}^{u}$, which are independent from the recommendations of the system.

\section{Evaluations}


\begin{table*}[t!h]
\centering
\caption{Comparison of accumulated regret under different environment settings.}
\vspace{-2mm}
\begin{tabular}{p{0.2cm} p{0.3cm} p{0.3cm} p{0.45cm} p{0.45cm} p{0.45cm} p{0.45cm} p{0.85cm} p{0.85cm} p{0.85cm} p{0.9cm} p{0.85cm} p{0.85cm} p{0.9cm}}
\toprule
& N & K & $S_{min}$ & $S_{max}$ & T & $\sigma$ & oracle. & LinUCB & adTS & dLinUCB & CLUB & SCLUB & CoDBand \\
\cmidrule(r{4pt}){2-7} \cmidrule(l){8-14}
1 & 100 & 10 & 500 & 3000 & 3000 & 0.1 & 124 & 24050 & 9183 & 3030 & 24602 & 24602 & \textbf{1193}\\
2 & 100 & 50 & 500 & 3000 & 3000 & 0.1 & 575 & 24352 & 19433 & 2858 & 24762 & 24980 & \textbf{2252} \\
3 & 100 & 100 & 500 & 3000 & 3000 & 0.1 & 922 & 28108 & 20828 & 3388 & 28424 & 28585 & \textbf{2688} \\
4 & 100 & 10 & 200 & 500 & 3000 & 0.1 & 128 & 54791 & 52282 & 17475 & 55098 & 55268 & \textbf{5143} \\
5 & 100 & 10 & 500 & 800 & 3000 & 0.1 & 131 & 51095 & 40538 & 8401 & 51440 & 51604 & \textbf{2423} \\
6 & 100 & 10 & 800 & 1100 & 3000 & 0.1 & 128 & 39035 & 26851 & 6549 & 39395 & 39477 & \textbf{2342} \\
7 & 100 & 10 & 500 & 3000 & 3000 & 0.13 & 175 & 27101 & 20555 & 3742 & 27163 & 27633 & \textbf{3043} \\
8 & 100 & 10 & 500 & 3000 & 3000 & 0.16 & 280 & 23949 & 21320 & 4833 & 23693 & 24436 & \textbf{3629} \\

\bottomrule
\end{tabular}
\label{tab:sim_table}
\vspace{-1mm}
\end{table*}

We performed extensive empirical evaluations of CoDBand against several related baseline bandit algorithms, which can be summarized into the following three categories. First, contextual bandits that do not consider collaboration effects or the non-stationarity of the environment: we include \textbf{LinUCB} \cite{LinUCB}, which has been shown to be effective in providing interactive personalized recommendations in a stationary environment. Second, collaborative bandits: 
\textbf{CLUB} \cite{gentile2014online}, which assumes the existence of underlying stationary user clusters and learns the user clusters and cluster-wise bandit models on the fly. \textbf{SCLUB} \cite{li2019improved}, which is a recent extension of \textbf{CLUB} for non-uniform distribution of the clusters.
Third, contextual bandits that account for a non-stationary environment in a per-user basis, including \textbf{AdTS} \cite{Hariri:2015:AUP:2832747.2832852}: which is an adaptive Thompson Sampling algorithm with a cumulative sum test based change detection module; and \textbf{dLinUCB}, which is a state-of-the-art non-stationary contextual bandit algorithm \cite{wu2018dLinUCB}. These two non-stationary solutions have shown to be the most competitive among the other non-stationary bandit solutions according to \cite{wu2018dLinUCB}. We compared all the algorithms in both simulations and large-scale real-world datasets to compare their effectiveness in handling a changing environment for collaborative recommendation. Our code for conducting these experiments will be available online.
In simulation-based experiments, we also include \textbf{oracle-LinUCB} for comparison, which runs an instance of LinUCB for each unique global bandit model in the corresponding stationary period in each user. Comparing with it helps us understand the added regret from errors in change detection and model clustering.
\subsection{Experiments on synthetic dataset}
\label{exp_simulation}
\noindent\textit{\textbf{Simulation settings:}}
In simulation, we generate a set of users $\mathcal{U}$ ($|\mathcal{U}| = N$) with an arm pool $\cA$ of size $1000$, in which each arm $a$ is associated with a $d$-dimensional feature vector $\bx \in \mathbb{R}^d$ with $\lVert \bx\rVert_2 \leq 1$. To simulate an abruptly changing environment, for each user we sample a sequence of time intervals from $(S_{min},S_{max})$ uniformly. Each time interval is considered as a stationary period such that we can naturally get the change points in each user. Note that since the stationary periods for different users are drawn independently, it is highly \emph{unlikely} for the users to change synchronously. 
At the change point $c$ of each user $u$, we experimented with three different settings to decide the ground-truth bandit parameters: 1) $\btheta_{u,c}$ is generated according to the DP model described in Eq \eqref{eq:CRP}; 2) $\btheta_{u,c}$ is sampled from a fixed set of unique bandit parameters $\left\{\phi_{k}\right\}_{k=1}^{K}$ with a predefined mixture weight; 3) a stationary environment is also included for comparison, where $\btheta_{u,c}$ remains the same over time.
Note that neither the users' change points, nor the ground-truth bandit parameters are disclosed to the learners. At each time step $t \in [T]$, all users in $\cU$ gets served one by one, and a subset of arms $\cA_{t} \subset \cA$ are randomly chosen and disclosed to the learner. The ground-truth reward $r_{t}$ is corrupted by Gaussian noise $\eta_{t} \sim N(0,\sigma^2)$ before giving back to the learners.

\noindent\textit{\textbf{Empirical regret comparison on synthetic dataset:}}
We set the number of user $N=100$, the total number of time steps $T=3000$, and the range of stationary period length for Settings 1 and 2 as $(S_{min}=500, S_{max}=3000)$. Setting 1 and 2 are initialized with the same set of unique bandit parameters of size $K=10$. We set $N=500$, $T=600$ and $K=2$ for setting 3. We report the accumulated regret of all algorithms under the three simulation settings in Figure \ref{fig:simulation}. We can observe that LinUCB, CLUB and SCLUB all suffer linear regret after the first change point in Setting 1 and 2 because of their strong but unrealistic stationary assumption. Both AdTS and dLinUCB can react to the environment changes, but they are slow and less accurate in doing so, and thus accumulate faster increasing regret. In addition, AdTS has a large probability of making false change detections and incurs fast increasing regret in the stationary Setting 3, where the underlying bandit model in each user does not change. The proposed CoDBand can not only quickly identify the changes in each user, but also properly recognize which existing model to reuse, which brings further reduction of regret comparing to those non-collaborative or non-stationary baselines. It is worth noting that in Setting 2, DP prior is mis-specified in CoDBand as the underlying bandit parameter generation does not follow this stochastic process, but CoDBand can still quickly identify the correct bandit model to use, and obtain better performance than all the baselines. In the three settings, the oracle-LinUCB baseline performed the best, as it knows exactly when the change happens and how the different users are related to each other. But the added regret from CoDBand is acceptable, given the algorithm needs to both detect the change and cluster the models on the fly without any prior knowledge about the environment.

To further verify the robustness of CoDBand under different simulation settings, we varied the parameters under Setting 2, e.g., the number of unique bandit parameters $K$, the minimum and maximum length for stationary periods $S_{min}$ and $S_{max}$, the standard deviation of noise $\sigma$, and report the result of algorithms' corresponding regret in Table \ref{tab:sim_table}. The results show that CoDBand can successfully cope with different environment settings and outperform the baselines. In addition, the trends of how regret changes with different parameters align with our regret analysis. For example, with the increase of the number of unique bandit parameter $K$ in the same number of $N$ users, the regret increases, because less observations can be shared among users. The regret also increases substantially with shorter stationary periods, as more errors would occur in change detection. In addition, larger amount of noise in the reward not only slows down CoDBand's bandit parameter estimation but also affects its change detection accuracy, and therefore leads to higher regret.
\vspace{-0.7mm}
\subsection{Experiments on real-world datasets}


\begin{figure*}[t]
\centering
\begin{tabular}{c c c c }
\includegraphics[width=4.1cm]{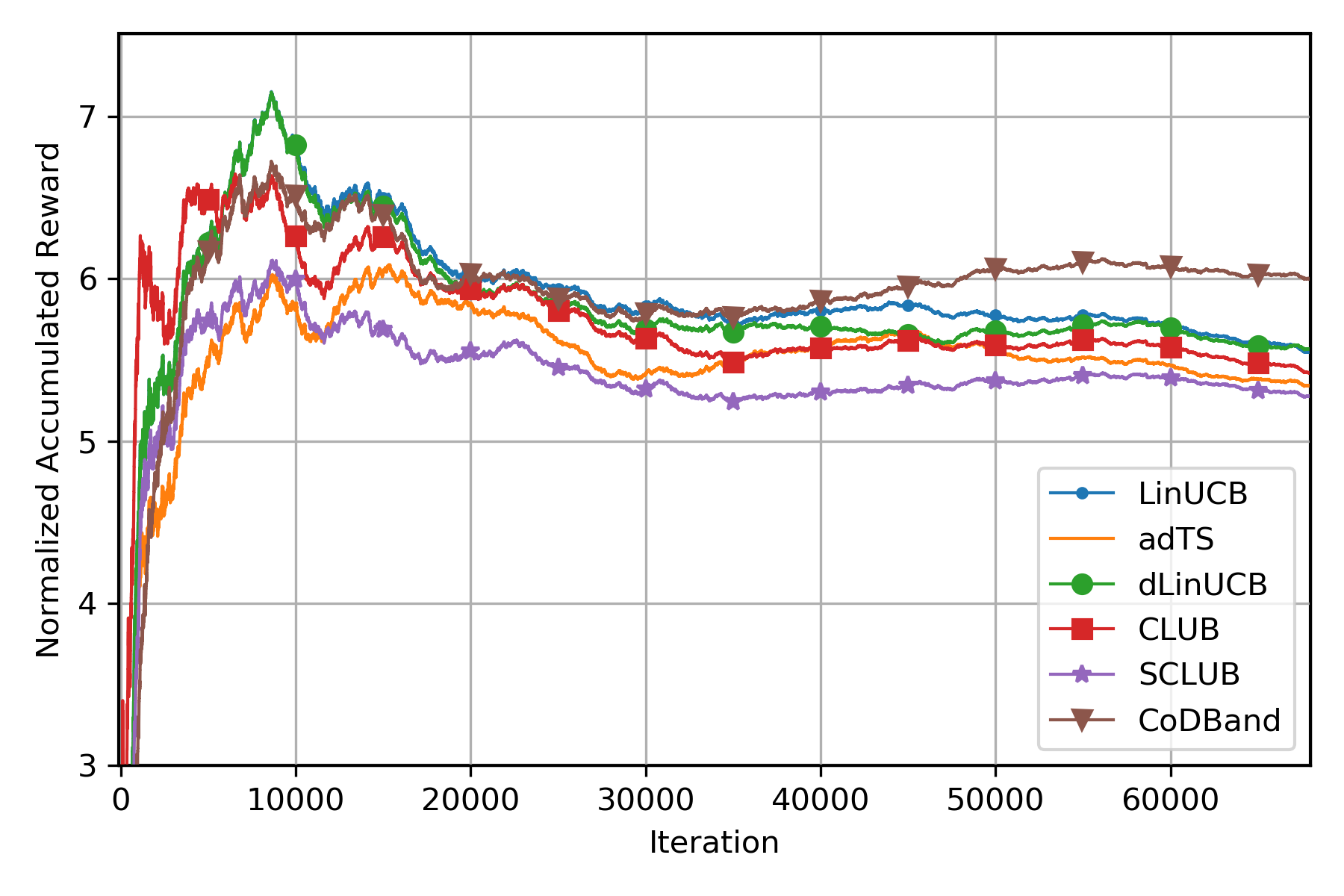} &
\includegraphics[width=4.1cm]{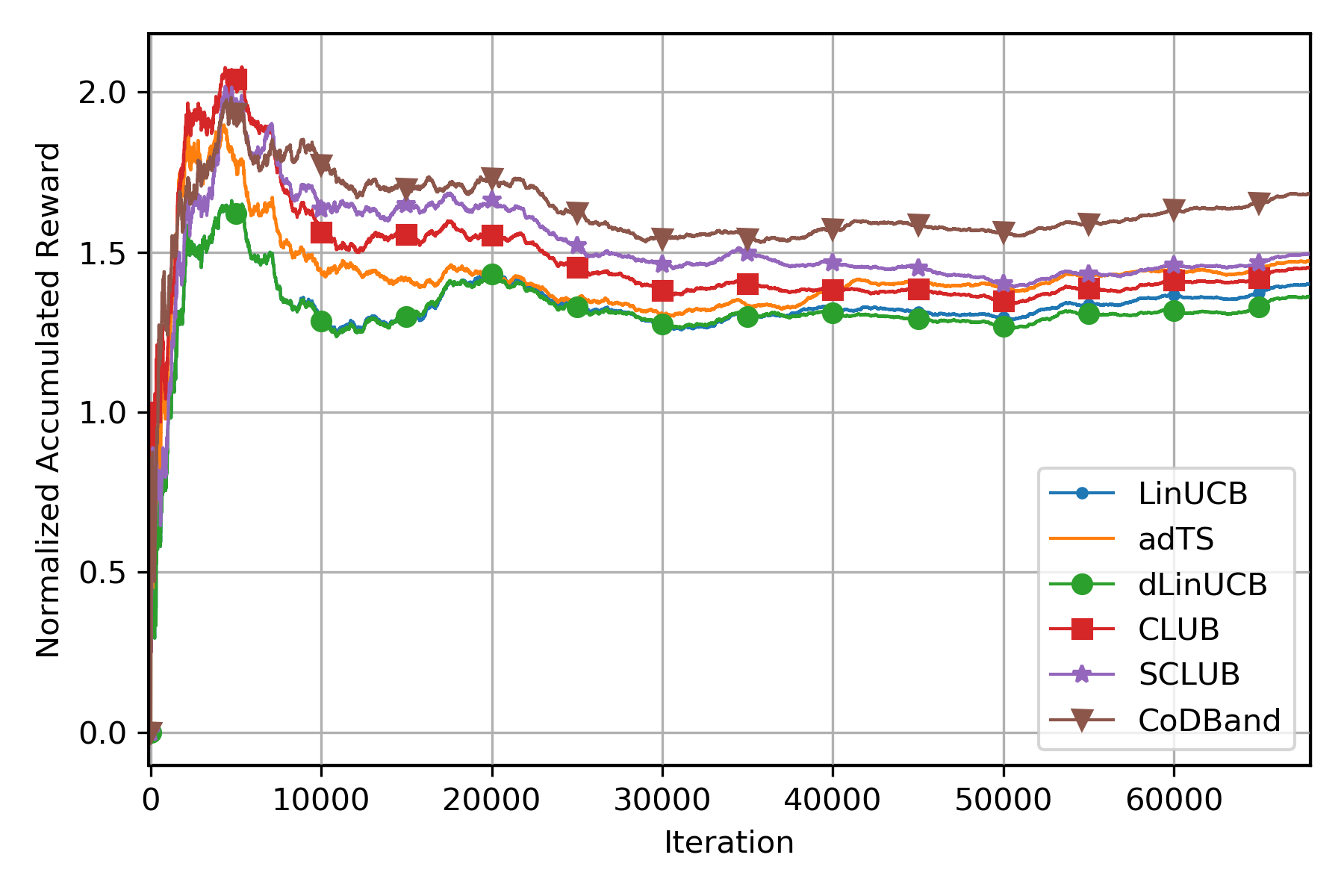} &
\includegraphics[width=4.1cm]{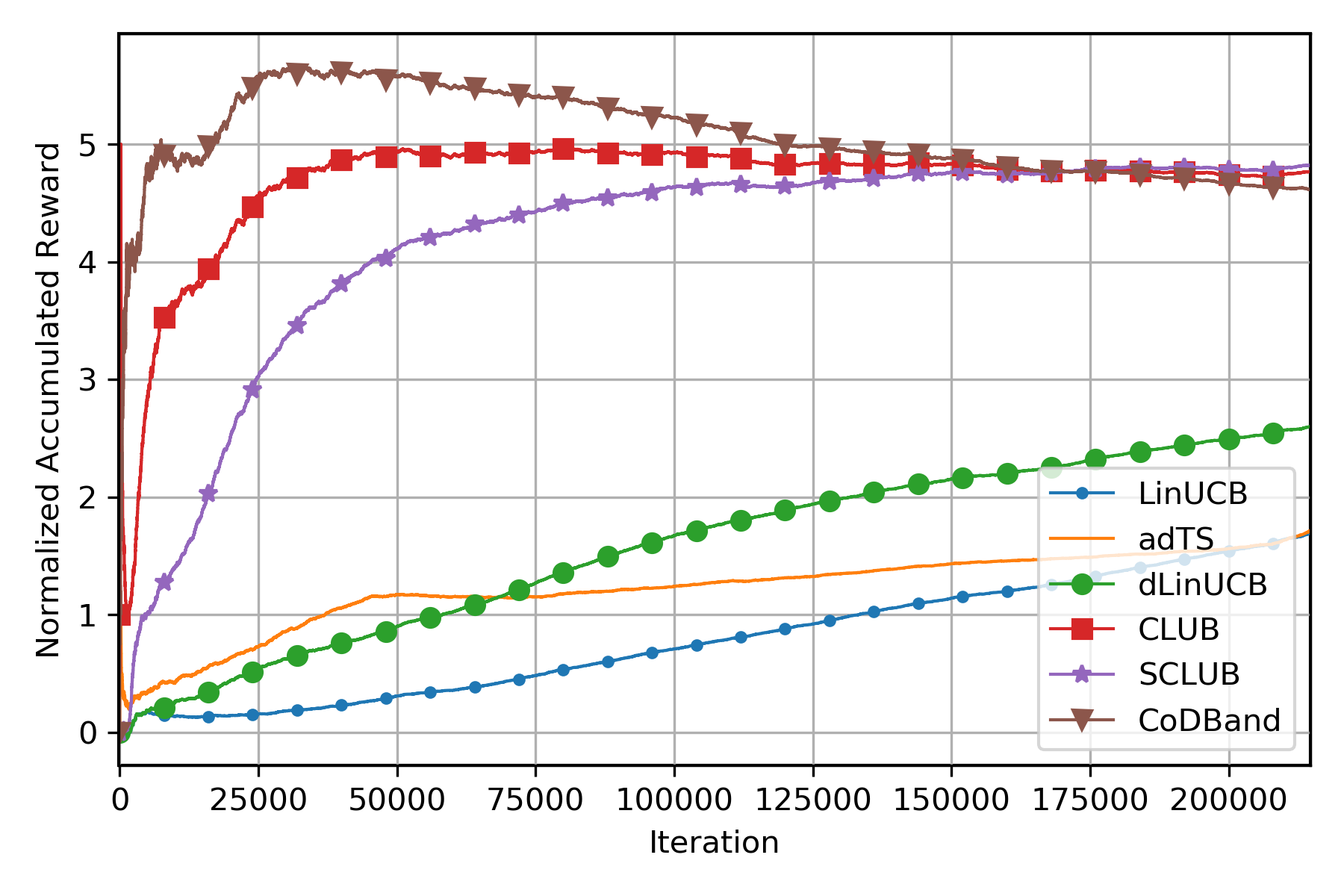} &
\includegraphics[width=4.1cm]{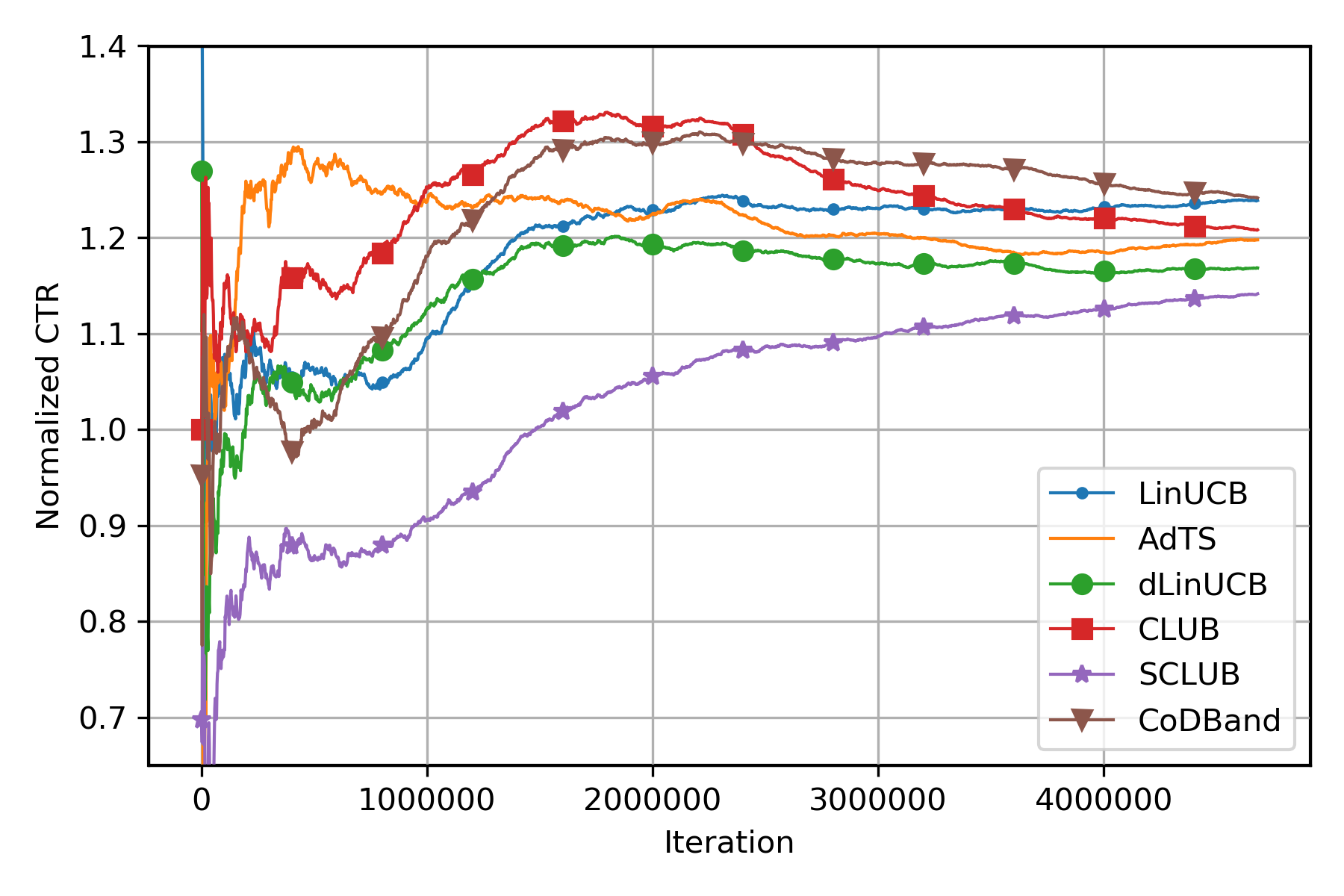} \\
\footnotesize (a)  Normalized reward on LastFM  & \footnotesize (b) Normalized reward on Delicious & \footnotesize (c) Normalized reward on MovieLens & \footnotesize (d) Normalized CTR on Yahoo! Today
\normalsize
\end{tabular}
\vspace{-1mm}
\caption{Performance comparison on realworld datasets.} \label{fig:real_world_reward}
\vspace{-1mm}
\end{figure*}


\noindent\textit{\textbf{LastFM and Delicious:}} The LastFM dataset is extracted from the music streaming service
Last.fm, and the Delicious dataset is extracted from the social bookmark sharing service Delicious. They were made availalbe by the HetRec 2011 workshop. 
The LastFM dataset contains 1892 users and 17632 items (artists). We consider the ``\textit{listened artists}'' in each user as positive feedback. The Delicious dataset contains 1861 users and 69226 items (URLs). We treat the bookmarked URLs in each user as positive feedback. 
Both datasets provide social network information about the users. 
Following the settings in \cite{Gang}, we pre-processed these two datasets in order to fit them into a contextual bandit setting. Firstly, we used all tags associated with an item to create a TF-IDF feature vector to represent each item. Then we used PCA to reduce the dimensionality of the feature vectors and retained the first 25 principle components to construct the context vectors, i.e., $d = 25$. We fixed the size of candidate arm pool to $|\cA_{t}|=25$; for a particular user $u$, we randomly picked one item from his/her nonzero reward items, and randomly picked the other 24 from those zero reward items. On these two datasets, since each individual user's observations are sparse and mostly collected from a short period of time, it is hard to directly observe non-stationarity. Previous studies \cite{wu2018dLinUCB,WSDM19_offline_evaluation_nonstationary} introduce non-stationarity in the following way: create 10 user groups (so-called super-user) via spectral clustering base on user social network. 
Users in the same user group are considered to have similar result preferences. Then the super-users are stacked together chronologically to create a hybrid user, i.e., non-stationarity. The boundaries between super-users are considered as preference change points of the hybrid user. In this work, to highlight the effectiveness of collaboration, we further make this non-stationary environment more challenging by splitting each super-user into 3 parts, and refer to them as mini-super users. We randomize the order of $3\times10$ mini-super users. In this case, collaborative bandit solutions should identify the overlap between mini-super users from the same super user and take advantage of observation sharing, while failing to detect such collaborative effects will cost an algorithm sub-optimal performance in such a setting. To clarify, in the rest of the discussions, when we mention ``user'' concerning LastFM and Delicious datasets, we are referring to the mini-super users.

We report normalized reward, e.g., the ratio between accumulative reward collected from the bandit algorithms and that from a random selection policy on LastFM and Delicious datasets in Figure \ref{fig:real_world_reward} (a) and (b) respectively. We can observe that on both datasets, CoDBand outperforms the baselines. 
The advantage of CoDBand is more apparent at the later stage of learning, where it accumulated enough observations to accurately estimate a set of global bandit models that were representative to predict result preferences of users in the population. These global bandit models can be used to provide high quality recommendations for new users or users that have recently switched their preferences, whereas the other baselines either got distracted by the outdated observations in their model estimation, or discarded the outdated observations and completely restart from scratch.

\begin{figure}[t]
\centering
\includegraphics[width=8cm]{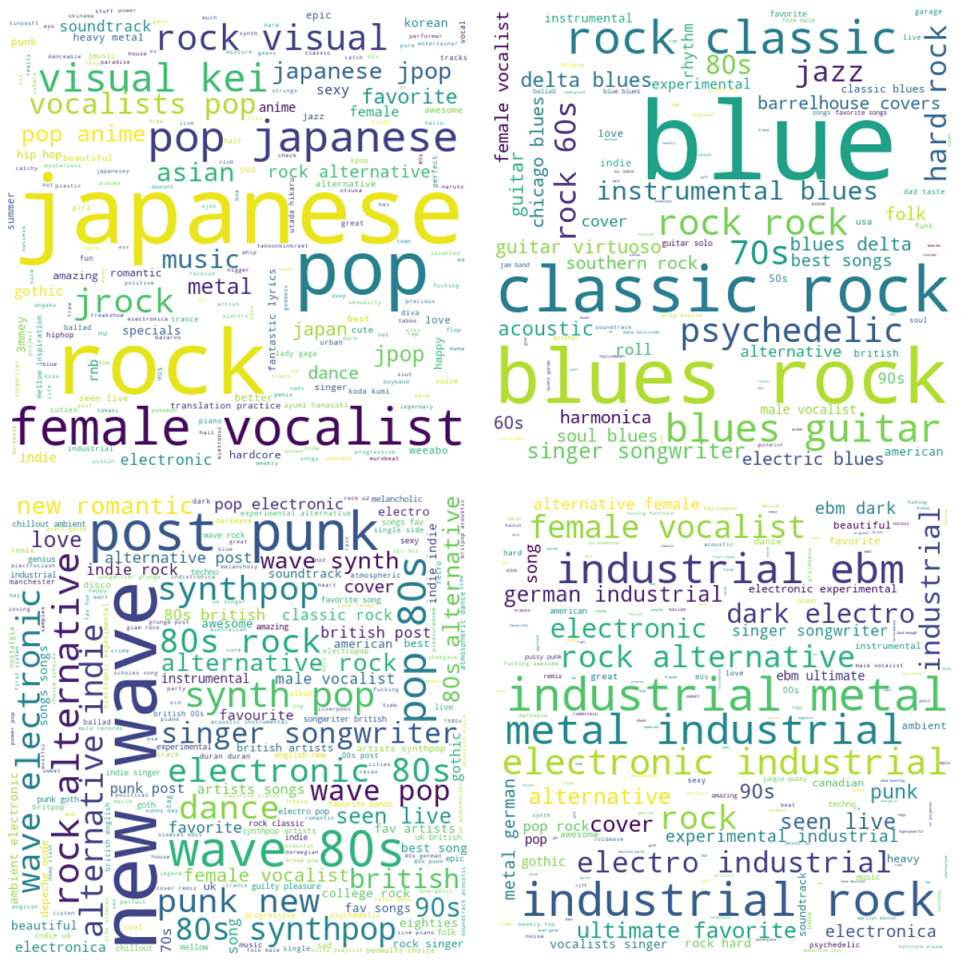}
\vspace{-2mm}
\caption{Word clouds for tags associated with the top-100 artists ranked by CoDBand's inferred most popular global bandit models.} \label{fig:word_cloud}
\vspace{-1mm}
\end{figure}

To further investigate what kind of result preferences in the user population that CoDBand has captured, we visualized its learnt global bandit models on the LastFM dataset. In this dataset, each artist is associated with a list of tags provided by the users. The tags are usually descriptive and reflect music genres or artist styles. For each global model learnt by CoDBand, we use the tags associated with the top-100 artists scored by this model to generate a word cloud. Figure \ref{fig:word_cloud} demonstrates four representative groups (based on their inferred popularity) CoDBand has learnt on LastFM, which clearly correspond to four different music genres –``\textit{J-pop}'', ``\textit{blues rock}'', ``\textit{new wave}'', and ``\textit{industrial metal}''. This qualitative result demonstrates CoDBand's capability in recognizing the potential clustering structure of users' preferences solely from their click feedback.

\noindent\textit{\textbf{MovieLens:}}
We also evaluated the algorithms with data extracted from the MovieLens 20M dataset that contains 20 million ratings with 27,000 movies and 138,000 users \cite{harper2015movielens}. We followed a similar procedure in \cite{li2016graph} to pre-process the data to fit a contextual bandit setting. 
First, we extracted TF-IDF feature vectors using information like movie titles, genres, and tags provided by users. We then applied PCA to the resulting TF-IDF feature vectors, and retained the first 25 principle components as the context vectors, i.e., $d=25$. Then we normalized all features to have a zero mean and unit variance. We converted ratings to binary reward by mapping non-zero ratings to 1, and zero ratings to 0. The event sequence is generated by first filtering out users with less than 3000 observations, and then at each time when a particular user $u$ is served, the candidate arm pool for user $u$ is generated by keeping the movie with nonzero reward at this time stamp and sampling another 24 zero-reward movies rated by this user, i.e., $|\cA_{t}|=25$.

We report the normalized accumulated reward of all algorithms in Figure \ref{fig:real_world_reward} (c). It is worth noticing that all the bandit algorithms with collaborative learning, e.g. CLUB, SCLUB and CoDBand perform substantially better than the other baselines. This indicates that users in the MovieLens dataset share much interests in common, and therefore data sharing is of vital importance for improving the performance. We can observe that CoDBand accumulated reward much faster than CLUB and SCLUB in the early stage. This suggests CoDBand is capable of estimating a good clustering structure over users with limited number of observations available and as a result starting to benefit from the shared observations earlier than CLUB and SCLUB. We attribute this advantage to its DP model based model selection solution, which leverages the concentration of user groups in a population of users (e.g., social norm). 
Though the non-stationary bandit algorithms dLinUCB and adTs also show improvement over standard LinUCB, not being able to utilize observations from other users make it hard for them to compete with the collaborative solutions on this dataset.




\noindent\textit{\textbf{Yahoo! Today Module:}}
Yahoo! Today Module recommendation dataset is a large-scale click stream dataset from the Yahoo Webscope program, which contains over 45 million user visits to Yahoo Today Module collected in 2009. For each visit, both the user and each of the 10 candidate articles, i.e. $|\cA_{t}|=10$, are associated with a feature vector of six dimensions ($d=5$ excluding a bias term) \cite{LinUCB}. We adopted the unbiased offline evaluation protocol in \cite{Li:2011:UOE:1935826.1935878} to compare the algorithms with data extracted from the first day of the ten-day period from this dataset, which contains 4.6 million user visits. Click through rate (CTR) is used as the performance metric for all bandit algorithms. Similar to \cite{LinUCB}, we normalized the resulting CTR of different algorithms by the corresponding logged random strategy’s CTR. In addition, this dataset does not provide user identities, we followed \cite{wu2016contextual,wu2018dLinUCB} to cluster users into different groups and view the resulting groups as users.

The results are reported in Figure \ref{fig:real_world_reward} (d). We can observe that CoDBand and CLUB show a faster and more steady rate in accumulating rewards than the other baselines, suggesting that considering collaboration among users is beneficial for this news recommendation scenario as well. While although AdTS exhibits faster increasing performance at the beginning, as it detects the changes in users' preference, its performance also deteriorates fast as it tends to make more false detections. It is also worth noticing that the simple baseline that attaches LinUCB to each individual user also performs reasonably, beating some of the other more complicated baselines. This suggests incorporating change detection or user clustering come with the risk of errors, e.g., false alarm in change detection causes the algorithm to discard observations when it is unnecessary, and including wrong user in the cluster introduces distortion to the learned model. These directly lead to the added regret comparing with standard baselines like LinUCB and SCLUB. On the other hand, the results in this experiment suggest CoDBand is more accurate in change detection and cluster identification, which ensures its advantage and flexibility against those more restrictive baselines. 





\section{Conclusions \& Future Work}
In this paper, we propose a collaborative dynamic bandit solution CoDBand for interactive recommendation in a non-stationary environment, where both user preferences and user dependencies can be changing over time. 
We model the changing environment with Dirichlet process, and propose a Thompson sampling based contextual bandit solution to perform collaborative online learning.
Rigorous regret analysis provides a valid performance guarantee of CoDBand for detecting the changes and correctly selecting the bandit models for recommendation. Extensive experiments on both synthetic and real-world datasets confirmed the effectiveness of CoDBand in recommendation, especially its advantages in helping addressing the cold start challenge. 

In our current formulation, the change points are assumed to happen at arbitrary and unknown time steps, and as a result they are outside of our Bayesian inference framework. A more elegant way is to also introduce a prior on the change points \citep{adams2007bayesian}, and use Thompson sampling to address both change detection and model selection \citep{mellor2013thompson}. Also in our current stochastic process model of the changing environment, we only explicitly modeled the popularity of bandit models with a Dirichlet Process model. But many other types of important observations can be considered, such as friendship and recency of a model. We would like further extend our Dirichlet Process model with other stochastic process models, such as Hawkes Process \cite{hawkes1974cluster}, to further enhance our solution in handling a complex changing environment.  

\begin{acks}
To Robert, for the bagels and explaining CMYK and color spaces.
\end{acks}

\bibliographystyle{ACM-Reference-Format}
\bibliography{bibfile}


\begin{thebibliography}{47}


\ifx \showCODEN    \undefined \def \showCODEN     #1{\unskip}     \fi
\ifx \showDOI      \undefined \def \showDOI       #1{#1}\fi
\ifx \showISBNx    \undefined \def \showISBNx     #1{\unskip}     \fi
\ifx \showISBNxiii \undefined \def \showISBNxiii  #1{\unskip}     \fi
\ifx \showISSN     \undefined \def \showISSN      #1{\unskip}     \fi
\ifx \showLCCN     \undefined \def \showLCCN      #1{\unskip}     \fi
\ifx \shownote     \undefined \def \shownote      #1{#1}          \fi
\ifx \showarticletitle \undefined \def \showarticletitle #1{#1}   \fi
\ifx \showURL      \undefined \def \showURL       {\relax}        \fi
\providecommand\bibfield[2]{#2}
\providecommand\bibinfo[2]{#2}
\providecommand\natexlab[1]{#1}
\providecommand\showeprint[2][]{arXiv:#2}

\bibitem[\protect\citeauthoryear{Abbasi-yadkori, P\'{a}l, and
  Szepesv\'{a}ri}{Abbasi-yadkori et~al\mbox{.}}{2011}]%
        {Improved_Algorithm}
\bibfield{author}{\bibinfo{person}{Yasin Abbasi-yadkori},
  \bibinfo{person}{D\'{a}vid P\'{a}l}, {and} \bibinfo{person}{Csaba
  Szepesv\'{a}ri}.} \bibinfo{year}{2011}\natexlab{}.
\newblock \showarticletitle{Improved Algorithms for Linear Stochastic Bandits}.
\newblock In \bibinfo{booktitle}{\emph{NIPS}}. \bibinfo{pages}{2312--2320}.
\newblock


\bibitem[\protect\citeauthoryear{Abeille and Lazaric}{Abeille and
  Lazaric}{2017}]%
        {pmlr-v54-abeille17a}
\bibfield{author}{\bibinfo{person}{Marc Abeille} {and}
  \bibinfo{person}{Alessandro Lazaric}.} \bibinfo{year}{2017}\natexlab{}.
\newblock \showarticletitle{{Linear Thompson Sampling Revisited}}. In
  \bibinfo{booktitle}{\emph{Proceedings of the 20th AISTATS}}.
  \bibinfo{pages}{176--184}.
\newblock


\bibitem[\protect\citeauthoryear{Adams and MacKay}{Adams and MacKay}{2007}]%
        {adams2007bayesian}
\bibfield{author}{\bibinfo{person}{Ryan~Prescott Adams} {and}
  \bibinfo{person}{David~JC MacKay}.} \bibinfo{year}{2007}\natexlab{}.
\newblock \showarticletitle{Bayesian online changepoint detection}.
\newblock \bibinfo{journal}{\emph{arXiv preprint arXiv:0710.3742}}
  (\bibinfo{year}{2007}).
\newblock


\bibitem[\protect\citeauthoryear{Agrawal and Goyal}{Agrawal and Goyal}{2013}]%
        {Agrawal:2013:TSC:3042817.3043073}
\bibfield{author}{\bibinfo{person}{Shipra Agrawal} {and} \bibinfo{person}{Navin
  Goyal}.} \bibinfo{year}{2013}\natexlab{}.
\newblock \showarticletitle{Thompson Sampling for Contextual Bandits with
  Linear Payoffs}. In \bibinfo{booktitle}{\emph{Proceedings of the 30th ICML}}.
  \bibinfo{pages}{1220--1228}.
\newblock


\bibitem[\protect\citeauthoryear{Antoniak}{Antoniak}{1974}]%
        {antoniak1974}
\bibfield{author}{\bibinfo{person}{Charles~E. Antoniak}.}
  \bibinfo{year}{1974}\natexlab{}.
\newblock \showarticletitle{Mixtures of Dirichlet Processes with Applications
  to Bayesian Nonparametric Problems}.
\newblock \bibinfo{journal}{\emph{The Annals of Statistics}}
  \bibinfo{volume}{2}, \bibinfo{number}{6} (\bibinfo{date}{11}
  \bibinfo{year}{1974}), \bibinfo{pages}{1152--1174}.
\newblock


\bibitem[\protect\citeauthoryear{Auer, Cesa-Bianchi, and Fischer}{Auer
  et~al\mbox{.}}{2002}]%
        {UCB1}
\bibfield{author}{\bibinfo{person}{Peter Auer}, \bibinfo{person}{Nicol\`{o}
  Cesa-Bianchi}, {and} \bibinfo{person}{Paul Fischer}.}
  \bibinfo{year}{2002}\natexlab{}.
\newblock \showarticletitle{Finite-time Analysis of the Multiarmed Bandit
  Problem}.
\newblock \bibinfo{journal}{\emph{Maching Learning}} \bibinfo{volume}{47},
  \bibinfo{number}{2-3} (\bibinfo{date}{May} \bibinfo{year}{2002}),
  \bibinfo{pages}{235--256}.
\newblock


\bibitem[\protect\citeauthoryear{Auer, Gajane, and Ortner}{Auer
  et~al\mbox{.}}{2019}]%
        {auer2019adaptively}
\bibfield{author}{\bibinfo{person}{Peter Auer}, \bibinfo{person}{Pratik
  Gajane}, {and} \bibinfo{person}{Ronald Ortner}.}
  \bibinfo{year}{2019}\natexlab{}.
\newblock \showarticletitle{Adaptively tracking the best bandit arm with an
  unknown number of distribution changes}. In
  \bibinfo{booktitle}{\emph{Conference on Learning Theory}}.
  \bibinfo{pages}{138--158}.
\newblock


\bibitem[\protect\citeauthoryear{Breese, Heckerman, and Kadie}{Breese
  et~al\mbox{.}}{1998}]%
        {breese1998empirical}
\bibfield{author}{\bibinfo{person}{John~S. Breese}, \bibinfo{person}{David
  Heckerman}, {and} \bibinfo{person}{Carl Kadie}.}
  \bibinfo{year}{1998}\natexlab{}.
\newblock \bibinfo{booktitle}{\emph{Empirical Analysis of Predictive Algorithms
  for Collaborative Filtering}}.
\newblock \bibinfo{type}{{T}echnical {R}eport} MSR-TR-98-12.
  \bibinfo{institution}{Microsoft Research}. \bibinfo{pages}{18} pages.
\newblock
\urldef\tempurl%
\url{http://research.microsoft.com/apps/pubs/default.aspx?id=69656}
\showURL{%
\tempurl}


\bibitem[\protect\citeauthoryear{Cesa-Bianchi, Gentile, and
  Zappella}{Cesa-Bianchi et~al\mbox{.}}{2013}]%
        {Gang}
\bibfield{author}{\bibinfo{person}{Nicolo Cesa-Bianchi},
  \bibinfo{person}{Claudio Gentile}, {and} \bibinfo{person}{Giovanni
  Zappella}.} \bibinfo{year}{2013}\natexlab{}.
\newblock \showarticletitle{A gang of bandits}.
\newblock  (\bibinfo{year}{2013}), \bibinfo{pages}{737--745}.
\newblock


\bibitem[\protect\citeauthoryear{Chapelle and Li}{Chapelle and Li}{2011}]%
        {chapelle2011empirical}
\bibfield{author}{\bibinfo{person}{Olivier Chapelle} {and}
  \bibinfo{person}{Lihong Li}.} \bibinfo{year}{2011}\natexlab{}.
\newblock \showarticletitle{An Empirical Evaluation of Thompson Sampling}. In
  \bibinfo{booktitle}{\emph{NIPS 2011}}. \bibinfo{pages}{2249--2257}.
\newblock


\bibitem[\protect\citeauthoryear{Chen, Lee, Luo, and Wei}{Chen
  et~al\mbox{.}}{2019}]%
        {chen2019new}
\bibfield{author}{\bibinfo{person}{Yifang Chen}, \bibinfo{person}{Chung-Wei
  Lee}, \bibinfo{person}{Haipeng Luo}, {and} \bibinfo{person}{Chen-Yu Wei}.}
  \bibinfo{year}{2019}\natexlab{}.
\newblock \showarticletitle{A new algorithm for non-stationary contextual
  bandits: Efficient, optimal, and parameter-free}.
\newblock \bibinfo{journal}{\emph{arXiv preprint arXiv:1902.00980}}
  (\bibinfo{year}{2019}).
\newblock


\bibitem[\protect\citeauthoryear{Cheung, Simchi-Levi, and Zhu}{Cheung
  et~al\mbox{.}}{2019}]%
        {cheung2019learning}
\bibfield{author}{\bibinfo{person}{Wang~Chi Cheung}, \bibinfo{person}{David
  Simchi-Levi}, {and} \bibinfo{person}{Ruihao Zhu}.}
  \bibinfo{year}{2019}\natexlab{}.
\newblock \showarticletitle{Learning to optimize under non-stationarity}. In
  \bibinfo{booktitle}{\emph{The 22nd AISTATS}}. \bibinfo{pages}{1079--1087}.
\newblock


\bibitem[\protect\citeauthoryear{Escobar and West}{Escobar and West}{1995}]%
        {escobar1995bayesian}
\bibfield{author}{\bibinfo{person}{Michael~D Escobar} {and}
  \bibinfo{person}{Mike West}.} \bibinfo{year}{1995}\natexlab{}.
\newblock \showarticletitle{Bayesian density estimation and inference using
  mixtures}.
\newblock \bibinfo{journal}{\emph{Journal of the american statistical
  association}} \bibinfo{volume}{90}, \bibinfo{number}{430}
  (\bibinfo{year}{1995}), \bibinfo{pages}{577--588}.
\newblock


\bibitem[\protect\citeauthoryear{Ferguson}{Ferguson}{1973}]%
        {ferguson1973}
\bibfield{author}{\bibinfo{person}{Thomas~S. Ferguson}.}
  \bibinfo{year}{1973}\natexlab{}.
\newblock \showarticletitle{A Bayesian Analysis of Some Nonparametric
  Problems}.
\newblock \bibinfo{journal}{\emph{The Annals of Statistics}}
  \bibinfo{volume}{1}, \bibinfo{number}{2} (\bibinfo{date}{03}
  \bibinfo{year}{1973}), \bibinfo{pages}{209--230}.
\newblock


\bibitem[\protect\citeauthoryear{Festinger}{Festinger}{1954}]%
        {doi:10.1177/001872675400700202}
\bibfield{author}{\bibinfo{person}{Leon Festinger}.}
  \bibinfo{year}{1954}\natexlab{}.
\newblock \showarticletitle{A Theory of Social Comparison Processes}.
\newblock \bibinfo{journal}{\emph{Human Relations}} \bibinfo{volume}{7},
  \bibinfo{number}{2} (\bibinfo{year}{1954}), \bibinfo{pages}{117--140}.
\newblock


\bibitem[\protect\citeauthoryear{Garivier and Moulines}{Garivier and
  Moulines}{[n.d.]}]%
        {garivier08_NonStationary}
\bibfield{author}{\bibinfo{person}{Aurélien Garivier} {and}
  \bibinfo{person}{Eric Moulines}.} \bibinfo{year}{[n.d.]}\natexlab{}.
\newblock \showarticletitle{On Upper-Confidence Bound Policies for
  Non-stationary Bandit Problems}. In \bibinfo{booktitle}{\emph{arXiv preprint
  arXiv:0805.3415 (2008)}}.
\newblock


\bibitem[\protect\citeauthoryear{Gentile, Li, Kar, Karatzoglou, Zappella, and
  Etrue}{Gentile et~al\mbox{.}}{2017}]%
        {pmlr-v70-gentile17a}
\bibfield{author}{\bibinfo{person}{Claudio Gentile}, \bibinfo{person}{Shuai
  Li}, \bibinfo{person}{Purushottam Kar}, \bibinfo{person}{Alexandros
  Karatzoglou}, \bibinfo{person}{Giovanni Zappella}, {and}
  \bibinfo{person}{Evans Etrue}.} \bibinfo{year}{2017}\natexlab{}.
\newblock \showarticletitle{On context-dependent clustering of bandits}. In
  \bibinfo{booktitle}{\emph{ICML}}. \bibinfo{pages}{1253--1262}.
\newblock


\bibitem[\protect\citeauthoryear{Gentile, Li, and Zappella}{Gentile
  et~al\mbox{.}}{2014}]%
        {gentile2014online}
\bibfield{author}{\bibinfo{person}{Claudio Gentile}, \bibinfo{person}{Shuai
  Li}, {and} \bibinfo{person}{Giovanni Zappella}.}
  \bibinfo{year}{2014}\natexlab{}.
\newblock \showarticletitle{Online Clustering of Bandits}. In
  \bibinfo{booktitle}{\emph{ICML'14}}. \bibinfo{pages}{757--765}.
\newblock


\bibitem[\protect\citeauthoryear{Hariri, Mobasher, and Burke}{Hariri
  et~al\mbox{.}}{2015}]%
        {Hariri:2015:AUP:2832747.2832852}
\bibfield{author}{\bibinfo{person}{Negar Hariri}, \bibinfo{person}{Bamshad
  Mobasher}, {and} \bibinfo{person}{Robin Burke}.}
  \bibinfo{year}{2015}\natexlab{}.
\newblock \showarticletitle{Adapting to user preference changes in interactive
  recommendation}. In \bibinfo{booktitle}{\emph{24th IJCAI}}.
\newblock


\bibitem[\protect\citeauthoryear{Harper and Konstan}{Harper and
  Konstan}{2015}]%
        {harper2015movielens}
\bibfield{author}{\bibinfo{person}{F~Maxwell Harper} {and}
  \bibinfo{person}{Joseph~A Konstan}.} \bibinfo{year}{2015}\natexlab{}.
\newblock \showarticletitle{The movielens datasets: History and context}.
\newblock \bibinfo{journal}{\emph{Acm transactions on interactive intelligent
  systems (tiis)}} \bibinfo{volume}{5}, \bibinfo{number}{4}
  (\bibinfo{year}{2015}), \bibinfo{pages}{1--19}.
\newblock


\bibitem[\protect\citeauthoryear{Hartland, Gelly, Baskiotis, Teytaud, and
  Sebag}{Hartland et~al\mbox{.}}{2006}]%
        {hartland:hal-00113668}
\bibfield{author}{\bibinfo{person}{Cedric Hartland}, \bibinfo{person}{Sylvain
  Gelly}, \bibinfo{person}{Nicolas Baskiotis}, \bibinfo{person}{Olivier
  Teytaud}, {and} \bibinfo{person}{Michele Sebag}.}
  \bibinfo{year}{2006}\natexlab{}.
\newblock \bibinfo{title}{{Multi-armed Bandit, Dynamic Environments and
  Meta-Bandits}}.  (\bibinfo{year}{2006}).
\newblock


\bibitem[\protect\citeauthoryear{Hawkes and Oakes}{Hawkes and Oakes}{1974}]%
        {hawkes1974cluster}
\bibfield{author}{\bibinfo{person}{Alan~G Hawkes} {and} \bibinfo{person}{David
  Oakes}.} \bibinfo{year}{1974}\natexlab{}.
\newblock \showarticletitle{A cluster process representation of a self-exciting
  process}.
\newblock \bibinfo{journal}{\emph{Journal of Applied Probability}}
  \bibinfo{volume}{11}, \bibinfo{number}{3} (\bibinfo{year}{1974}),
  \bibinfo{pages}{493--503}.
\newblock


\bibitem[\protect\citeauthoryear{Jagerman, Markov, and Rijke}{Jagerman
  et~al\mbox{.}}{2019}]%
        {WSDM19_offline_evaluation_nonstationary}
\bibfield{author}{\bibinfo{person}{Rolf Jagerman}, \bibinfo{person}{Ilya
  Markov}, {and} \bibinfo{person}{Maarten~de Rijke}.}
  \bibinfo{year}{2019}\natexlab{}.
\newblock \showarticletitle{When People Change their Mind: Off-Policy
  Evaluation in Non-stationary Recommendation Environments}. In
  \bibinfo{booktitle}{\emph{Proceedings of 12th WSDM}}. ACM,
  \bibinfo{pages}{297--306}.
\newblock


\bibitem[\protect\citeauthoryear{Kawale, Bui, Kveton, Tran-Thanh, and
  Chawla}{Kawale et~al\mbox{.}}{2015}]%
        {kawale2015efficient}
\bibfield{author}{\bibinfo{person}{Jaya Kawale}, \bibinfo{person}{Hung~H Bui},
  \bibinfo{person}{Branislav Kveton}, \bibinfo{person}{Long Tran-Thanh}, {and}
  \bibinfo{person}{Sanjay Chawla}.} \bibinfo{year}{2015}\natexlab{}.
\newblock \showarticletitle{Efficient Thompson Sampling for Online
  Matrix-Factorization Recommendation}. In \bibinfo{booktitle}{\emph{NIPS}}.
  \bibinfo{pages}{1297--1305}.
\newblock


\bibitem[\protect\citeauthoryear{Koren, Bell, and Volinsky}{Koren
  et~al\mbox{.}}{2009}]%
        {koren2009matrix}
\bibfield{author}{\bibinfo{person}{Yehuda Koren}, \bibinfo{person}{Robert
  Bell}, {and} \bibinfo{person}{Chris Volinsky}.}
  \bibinfo{year}{2009}\natexlab{}.
\newblock \showarticletitle{Matrix factorization techniques for recommender
  systems}.
\newblock \bibinfo{journal}{\emph{Computer}} \bibinfo{number}{8}
  (\bibinfo{year}{2009}), \bibinfo{pages}{30--37}.
\newblock


\bibitem[\protect\citeauthoryear{Lattimore and Szepesv{\'a}ri}{Lattimore and
  Szepesv{\'a}ri}{2020}]%
        {lattimore2020bandit}
\bibfield{author}{\bibinfo{person}{Tor Lattimore} {and} \bibinfo{person}{Csaba
  Szepesv{\'a}ri}.} \bibinfo{year}{2020}\natexlab{}.
\newblock \bibinfo{booktitle}{\emph{Bandit algorithms}}.
\newblock \bibinfo{publisher}{Cambridge University Press}.
\newblock


\bibitem[\protect\citeauthoryear{Li, Chu, Langford, and Schapire}{Li
  et~al\mbox{.}}{2010}]%
        {LinUCB}
\bibfield{author}{\bibinfo{person}{Lihong Li}, \bibinfo{person}{Wei Chu},
  \bibinfo{person}{John Langford}, {and} \bibinfo{person}{Robert~E Schapire}.}
  \bibinfo{year}{2010}\natexlab{}.
\newblock \showarticletitle{A contextual-bandit approach to personalized news
  article recommendation}. In \bibinfo{booktitle}{\emph{Proceedings of 19th
  WWW}}. ACM, \bibinfo{pages}{661--670}.
\newblock


\bibitem[\protect\citeauthoryear{Li, Chu, Langford, and Wang}{Li
  et~al\mbox{.}}{2011}]%
        {Li:2011:UOE:1935826.1935878}
\bibfield{author}{\bibinfo{person}{Lihong Li}, \bibinfo{person}{Wei Chu},
  \bibinfo{person}{John Langford}, {and} \bibinfo{person}{Xuanhui Wang}.}
  \bibinfo{year}{2011}\natexlab{}.
\newblock \showarticletitle{Unbiased offline evaluation of
  contextual-bandit-based news article recommendation algorithms}. In
  \bibinfo{booktitle}{\emph{Proceedings of 4th WSDM}}. ACM,
  \bibinfo{pages}{297--306}.
\newblock


\bibitem[\protect\citeauthoryear{Li, Chen, and Leung}{Li et~al\mbox{.}}{2019}]%
        {li2019improved}
\bibfield{author}{\bibinfo{person}{Shuai Li}, \bibinfo{person}{Wei Chen}, {and}
  \bibinfo{person}{Kwong-Sak Leung}.} \bibinfo{year}{2019}\natexlab{}.
\newblock \showarticletitle{Improved algorithm on online clustering of
  bandits}.
\newblock \bibinfo{journal}{\emph{arXiv preprint arXiv:1902.09162}}
  (\bibinfo{year}{2019}).
\newblock


\bibitem[\protect\citeauthoryear{Li, Gentile, and Karatzoglou}{Li
  et~al\mbox{.}}{2016a}]%
        {li2016graph}
\bibfield{author}{\bibinfo{person}{Shuai Li}, \bibinfo{person}{Claudio
  Gentile}, {and} \bibinfo{person}{Alexandros Karatzoglou}.}
  \bibinfo{year}{2016}\natexlab{a}.
\newblock \showarticletitle{Graph clustering bandits for recommendation}.
\newblock \bibinfo{journal}{\emph{arXiv preprint arXiv:1605.00596}}
  (\bibinfo{year}{2016}).
\newblock


\bibitem[\protect\citeauthoryear{Li, Karatzoglou, and Gentile}{Li
  et~al\mbox{.}}{2016b}]%
        {Li:2016:CFB:2911451.2911548}
\bibfield{author}{\bibinfo{person}{Shuai Li}, \bibinfo{person}{Alexandros
  Karatzoglou}, {and} \bibinfo{person}{Claudio Gentile}.}
  \bibinfo{year}{2016}\natexlab{b}.
\newblock \showarticletitle{Collaborative Filtering Bandits}. In
  \bibinfo{booktitle}{\emph{Proceedings of the 39th ACM SIGIR}}.
  \bibinfo{pages}{539--548}.
\newblock


\bibitem[\protect\citeauthoryear{Luo, Wei, Agarwal, and Langford}{Luo
  et~al\mbox{.}}{2018}]%
        {luo2018efficient}
\bibfield{author}{\bibinfo{person}{Haipeng Luo}, \bibinfo{person}{Chen-Yu Wei},
  \bibinfo{person}{Alekh Agarwal}, {and} \bibinfo{person}{John Langford}.}
  \bibinfo{year}{2018}\natexlab{}.
\newblock \showarticletitle{Efficient contextual bandits in non-stationary
  worlds}. In \bibinfo{booktitle}{\emph{Conference On Learning Theory}}.
  \bibinfo{pages}{1739--1776}.
\newblock


\bibitem[\protect\citeauthoryear{Mellor and Shapiro}{Mellor and
  Shapiro}{2013}]%
        {mellor2013thompson}
\bibfield{author}{\bibinfo{person}{Joseph Mellor} {and}
  \bibinfo{person}{Jonathan Shapiro}.} \bibinfo{year}{2013}\natexlab{}.
\newblock \showarticletitle{Thompson sampling in switching environments with
  bayesian online change point detection}.
\newblock \bibinfo{journal}{\emph{arXiv preprint arXiv:1302.3721}}
  (\bibinfo{year}{2013}).
\newblock


\bibitem[\protect\citeauthoryear{Moore, Chen, Turnbull, and Joachims}{Moore
  et~al\mbox{.}}{2013}]%
        {Moore2013TasteOT}
\bibfield{author}{\bibinfo{person}{Joshua~L Moore}, \bibinfo{person}{Shuo
  Chen}, \bibinfo{person}{Douglas Turnbull}, {and} \bibinfo{person}{Thorsten
  Joachims}.} \bibinfo{year}{2013}\natexlab{}.
\newblock \showarticletitle{Taste Over Time: The Temporal Dynamics of User
  Preferences.}. In \bibinfo{booktitle}{\emph{ISMIR}}.
  \bibinfo{pages}{401--406}.
\newblock


\bibitem[\protect\citeauthoryear{Radinsky, Svore, Dumais, Teevan, Bocharov, and
  Horvitz}{Radinsky et~al\mbox{.}}{2012}]%
        {Radinsky:2012:MPB:2187836.2187918}
\bibfield{author}{\bibinfo{person}{Kira Radinsky}, \bibinfo{person}{Krysta
  Svore}, \bibinfo{person}{Susan Dumais}, \bibinfo{person}{Jaime Teevan},
  \bibinfo{person}{Alex Bocharov}, {and} \bibinfo{person}{Eric Horvitz}.}
  \bibinfo{year}{2012}\natexlab{}.
\newblock \showarticletitle{Modeling and predicting behavioral dynamics on the
  web}. In \bibinfo{booktitle}{\emph{Proceedings of the 21st international
  conference on World Wide Web}}. \bibinfo{pages}{599--608}.
\newblock


\bibitem[\protect\citeauthoryear{Resnick and Varian}{Resnick and
  Varian}{1997}]%
        {resnick1997recommender}
\bibfield{author}{\bibinfo{person}{Paul Resnick} {and} \bibinfo{person}{Hal~R
  Varian}.} \bibinfo{year}{1997}\natexlab{}.
\newblock \showarticletitle{Recommender systems}.
\newblock \bibinfo{journal}{\emph{Commun. ACM}} \bibinfo{volume}{40},
  \bibinfo{number}{3} (\bibinfo{year}{1997}), \bibinfo{pages}{56--58}.
\newblock


\bibitem[\protect\citeauthoryear{Russac, Vernade, and Capp{\'e}}{Russac
  et~al\mbox{.}}{2019}]%
        {russac2019weighted}
\bibfield{author}{\bibinfo{person}{Yoan Russac}, \bibinfo{person}{Claire
  Vernade}, {and} \bibinfo{person}{Olivier Capp{\'e}}.}
  \bibinfo{year}{2019}\natexlab{}.
\newblock \showarticletitle{Weighted Linear Bandits for Non-Stationary
  Environments}. In \bibinfo{booktitle}{\emph{NIPS}}.
  \bibinfo{pages}{12017--12026}.
\newblock


\bibitem[\protect\citeauthoryear{Russo and Van~Roy}{Russo and Van~Roy}{2014}]%
        {russo2014learning}
\bibfield{author}{\bibinfo{person}{Daniel Russo} {and}
  \bibinfo{person}{Benjamin Van~Roy}.} \bibinfo{year}{2014}\natexlab{}.
\newblock \showarticletitle{Learning to optimize via posterior sampling}.
\newblock \bibinfo{journal}{\emph{Mathematics of Operations Research}}
  \bibinfo{volume}{39}, \bibinfo{number}{4} (\bibinfo{year}{2014}),
  \bibinfo{pages}{1221--1243}.
\newblock


\bibitem[\protect\citeauthoryear{Sarwar, Karypis, Konstan, and Riedl}{Sarwar
  et~al\mbox{.}}{2001}]%
        {sarwar2001item}
\bibfield{author}{\bibinfo{person}{Badrul Sarwar}, \bibinfo{person}{George
  Karypis}, \bibinfo{person}{Joseph Konstan}, {and} \bibinfo{person}{John
  Riedl}.} \bibinfo{year}{2001}\natexlab{}.
\newblock \showarticletitle{Item-based collaborative filtering recommendation
  algorithms}. In \bibinfo{booktitle}{\emph{Proceedings of 10th WWW}}. ACM,
  \bibinfo{pages}{285--295}.
\newblock


\bibitem[\protect\citeauthoryear{Tantipathananandh, Berger-Wolf, and
  Kempe}{Tantipathananandh et~al\mbox{.}}{2007}]%
        {tantipathananandh2007framework}
\bibfield{author}{\bibinfo{person}{Chayant Tantipathananandh},
  \bibinfo{person}{Tanya Berger-Wolf}, {and} \bibinfo{person}{David Kempe}.}
  \bibinfo{year}{2007}\natexlab{}.
\newblock \showarticletitle{A framework for community identification in dynamic
  social networks}. In \bibinfo{booktitle}{\emph{Proceedings of the 13th ACM
  KDD}}. ACM, \bibinfo{pages}{717--726}.
\newblock


\bibitem[\protect\citeauthoryear{Wang, Wu, and Wang}{Wang
  et~al\mbox{.}}{2017}]%
        {FactorUCB}
\bibfield{author}{\bibinfo{person}{Huazheng Wang}, \bibinfo{person}{Qingyun
  Wu}, {and} \bibinfo{person}{Hongning Wang}.} \bibinfo{year}{2017}\natexlab{}.
\newblock \showarticletitle{Factorization Bandits for Interactive
  Recommendation}. In \bibinfo{booktitle}{\emph{AAAI}}.
  \bibinfo{pages}{2695--2702}.
\newblock


\bibitem[\protect\citeauthoryear{Wu, Iyer, and Wang}{Wu et~al\mbox{.}}{2018}]%
        {wu2018dLinUCB}
\bibfield{author}{\bibinfo{person}{Qingyun Wu}, \bibinfo{person}{Naveen Iyer},
  {and} \bibinfo{person}{Hongning Wang}.} \bibinfo{year}{2018}\natexlab{}.
\newblock \showarticletitle{Learning contextual bandits in a non-stationary
  environment}. In \bibinfo{booktitle}{\emph{The 41st International ACM
  SIGIR}}. ACM, \bibinfo{pages}{495--504}.
\newblock


\bibitem[\protect\citeauthoryear{Wu, Wang, Gu, and Wang}{Wu
  et~al\mbox{.}}{2016}]%
        {wu2016contextual}
\bibfield{author}{\bibinfo{person}{Qingyun Wu}, \bibinfo{person}{Huazheng
  Wang}, \bibinfo{person}{Quanquan Gu}, {and} \bibinfo{person}{Hongning Wang}.}
  \bibinfo{year}{2016}\natexlab{}.
\newblock \showarticletitle{Contextual Bandits in a Collaborative Environment}.
  In \bibinfo{booktitle}{\emph{Proceedings of the 39th International ACM
  SIGIR}}. ACM, \bibinfo{pages}{529--538}.
\newblock


\bibitem[\protect\citeauthoryear{Wu, Wang, Hong, and Shi}{Wu
  et~al\mbox{.}}{2017}]%
        {wu2017returning}
\bibfield{author}{\bibinfo{person}{Qingyun Wu}, \bibinfo{person}{Hongning
  Wang}, \bibinfo{person}{Liangjie Hong}, {and} \bibinfo{person}{Yue Shi}.}
  \bibinfo{year}{2017}\natexlab{}.
\newblock \showarticletitle{Returning is believing: Optimizing long-term user
  engagement in recommender systems}. In \bibinfo{booktitle}{\emph{Proceedings
  of the 26th ACM CIKM}}. ACM, \bibinfo{pages}{1927--1936}.
\newblock


\bibitem[\protect\citeauthoryear{Yang, Toni, and Dong}{Yang
  et~al\mbox{.}}{2020}]%
        {yang2020laplacian}
\bibfield{author}{\bibinfo{person}{Kaige Yang}, \bibinfo{person}{Laura Toni},
  {and} \bibinfo{person}{Xiaowen Dong}.} \bibinfo{year}{2020}\natexlab{}.
\newblock \showarticletitle{Laplacian-regularized graph bandits: Algorithms and
  theoretical analysis}. In \bibinfo{booktitle}{\emph{AISTATS}}.
  \bibinfo{pages}{3133--3143}.
\newblock


\bibitem[\protect\citeauthoryear{Yu and Mannor}{Yu and Mannor}{2009}]%
        {yu2009piecewise}
\bibfield{author}{\bibinfo{person}{Jia~Yuan Yu} {and} \bibinfo{person}{Shie
  Mannor}.} \bibinfo{year}{2009}\natexlab{}.
\newblock \showarticletitle{Piecewise-stationary bandit problems with side
  observations}. In \bibinfo{booktitle}{\emph{Proceedings of the 26th ICML}}.
  ACM, \bibinfo{pages}{1177--1184}.
\newblock


\bibitem[\protect\citeauthoryear{Zhao, Zhang, Jiang, and Zhou}{Zhao
  et~al\mbox{.}}{2020}]%
        {zhao2020simple}
\bibfield{author}{\bibinfo{person}{Peng Zhao}, \bibinfo{person}{Lijun Zhang},
  \bibinfo{person}{Yuan Jiang}, {and} \bibinfo{person}{Zhi-Hua Zhou}.}
  \bibinfo{year}{2020}\natexlab{}.
\newblock \showarticletitle{A simple approach for non-stationary linear
  bandits}. In \bibinfo{booktitle}{\emph{Proceedings of the 23rd AISTATS}},
  Vol.~\bibinfo{volume}{2020}.
\newblock


\end{thebibliography}


\section{Proof of Lemma \ref{lem:regret_decomposition}}
We can decompose the instantaneous Bayesian regret at time $t$ as:
\small
\begin{equation*}
\begin{split}
\bbE[r_{t}] &= \bbE[f_{\theta_{i_{t},t}}(\bx_{t}^{*})-f_{\theta_{i_{t},t}}(\bx_{t})] = \bbE[\bbE[f_{\theta_{i_{t},t}}(\bx_{t}^{*})-f_{\theta_{i_{t},t}}(\bx_{t})|\cH_{t-1}]] \\
& = \bbE[\bbE[U_{t}(\tilde{z}_{t}, \bx_{t})-U_{t}(z^{*}_{t},\bx_{t}^{*})+f_{\theta_{i_{t},t}}(\bx_{t}^{*})-f_{\theta_{i_{t},t}}(\bx_{t})|\cH_{t-1}]] \\
& = \bbE[U_{t}(\tilde{z}_{t}, \bx_{t})-f_{\theta_{i_{t},t}}(\bx_{t})] + \bbE[f_{\theta_{i_{t},t}}(\bx_{t}^{*}) - U_{t}(z^{*}_{t},\bx_{t}^{*})]
\end{split}
\end{equation*}
\normalsize
Then we can decompose the first term $\bbE[U_{t}(\tilde{z}_{t}, \bx_{t})-f_{\theta_{i_{t},t}}(\bx_{t})]$ into:
\small
\begin{equation*}
\begin{split}
    & \bbE[U_{t}(\tilde{z}_{t}, \bx_{t})-f_{\theta_{i_{t},t}}(\bx_{t})] \\
    & = \bbE[U_{t}(\tilde{z}_{t}, \bx_{t})-U_{t}(z^{*}_{t}, \bx_{t})+U_{t}(z^{*}_{t}, \bx_{t})-f_{\theta_{i_{t},t}}(\bx_{t})] \\
    & = \bbE[U_{t}(\tilde{z}_{t}, \bx_{t})-U_{t}(z^{*}_{t}, \bx_{t})]+ \bbE[U_{t}(z^{*}_{t}, \bx_{t})-f_{\theta_{i_{t},t}}(\bx_{t})] \\
    & = \bbE[\bigl[U_{t}(\tilde{z}_{t}, \bx_{t})-U_{t}(z^{*}_{t}, \bx_{t})\bigr]\cdot \mathbf{1}\bigl\{\tilde{z}_{t}=z^{*}_{t}\bigr\}] \\
    & + \bbE[\bigl[U_{t}(\tilde{z}_{t}, \bx_{t})-U_{t}(z^{*}_{t}, \bx_{t})\bigr]\cdot \mathbf{1}\bigl\{\tilde{z}_{t} \neq z^{*}_{t}\bigr\}]+ \bbE[U_{t}(z^{*}_{t}, \bx_{t})-f_{\theta_{i_{t},t}}(\bx_{t})] \\
    & = \bbE[\bigl[U_{t}(\tilde{z}_{t}, \bx_{t})-U_{t}(z^{*}_{t}, \bx_{t})\bigr]\cdot \mathbf{1}\bigl\{\tilde{z}_{t} \neq z^{*}_{t}\bigr\}]+ \bbE[U_{t}(z^{*}_{t}, \bx_{t})-f_{\theta_{i_{t},t}}(\bx_{t})] \\
\end{split}
\end{equation*}
\normalsize
And $\bbE[U_{t}(z^{*}_{t}, \bx_{t})-f_{\theta_{i_{t},t}}(\bx_{t})]$ can be upper bounded by:
\small
\begin{equation*}
\begin{split}
    & \bbE[[U_{t}(z^{*}_{t}, \bx_{t})-f_{\theta_{i_{t},t}}(\bx_{t})] \cdot \mathbf{1}\{f_{\theta_{i_{t},t}}(\bx_{t}) \geq L_{t}(z^{*}_{t}, \bx_{t})\} \\
    & + [U_{t}(z^{*}_{t},\bx_{t})-f_{\theta_{i_{t},t}}(\bx_{t})] \cdot \mathbf{1}\{f_{\theta_{i_{t},t}}(\bx_{t}) < L_{t}(z^{*}_{t}, \bx_{t})\}] \\
    & \leq \bbE[U_{t}(z^{*}_{t}, \bx_{t})-L_{t}(z^{*}_{t}, \bx_{t})] \cdot \bbP\{f_{\theta_{i_{t},t}}(\bx_{t}) \geq L_{t}(z^{*}_{t}, \bx_{t})\} \\
    & + 2 \cdot \bbP\{f_{\theta_{i_{t},t}}(\bx_{t}) < L_{t}(z^{*}_{t}, \bx_{t})\} \\
    & \leq \bbE[U_{t}(z^{*}_{t}, \bx_{t})-L_{t}(z^{*}_{t}, \bx_{t})] + 2 \cdot \bbP\{f_{\theta_{i_{t},t}}(\bx_{t}) < L_{t}(z^{*}_{t}, \bx_{t})\} \\
\end{split}
\end{equation*}
\normalsize
The second term $\bbE[f_{\theta_{i_{t},t}}(\bx_{t}^{*}) - U_{t}(z^{*}_{t},\bx_{t}^{*})]$ can be decomposed into:
\begin{equation*}
\begin{split}
& \bbE[[f_{\theta_{i_{t},t}}(\bx_{t}^{*})-U_{t}(z^{*}_{t},\bx_{t}^{*})] \cdot \mathbf{1}\{f_{\theta_{i_{t},t}}(\bx_{t}^{*}) \leq U_{t}(z^{*}_{t},\bx_{t}^{*})\} \\
& + [f_{\theta_{i_{t},t}}(\bx_{t}^{*})-U_{t}(z^{*}_{t},\bx_{t}^{*})] \cdot \mathbf{1}\{f_{\theta_{i_{t},t}}(\bx_{t}^{*}) > U_{t}(z^{*}_{t},\bx_{t}^{*})\}] \\
& \leq 0 \cdot \bbP\{f_{\theta_{i_{t},t}}(\bx_{t}^{*}) \leq U_{t}(z^{*}_{t},\bx_{t}^{*})\} + 2 \cdot \bbP\{f_{\theta_{i_{t},t}}(\bx_{t}^{*}) > U_{t}(z^{*}_{t},\bx_{t}^{*})\} \\
& = 2 \cdot \bbP\{f_{\theta_{i_{t},t}}(\bx_{t}^{*}) > U_{t}(z^{*}_{t},\bx_{t}^{*})\} \\
\end{split}
\end{equation*}

Combining everything we have the following upper bound on instantaneous Bayesian regret:
\small
\begin{equation*}
\begin{split}
&\bbE[r_{t}] \\
& \leq \bbE[\bigl[U_{t}(\tilde{z}_{t}, \bx_{t})-U_{t}(z^{*}_{t}, \bx_{t})\bigr]\cdot \mathbf{1}\bigl\{\tilde{z}_{t} \neq z^{*}_{t}\bigr\}]+ \bbE[U_{t}(z^{*}_{t}, \bx_{t})-L_{t}(z^{*}_{t}, \bx_{t})] \\
& + 2 \cdot \bbP\{f_{\theta_{i_{t},t}}(\bx_{t}) < L_{t}(z^{*}_{t}, \bx_{t})\} + 2 \cdot \bbP\{f_{\theta_{i_{t},t}}(\bx_{t}^{*}) > U_{t}(z^{*}_{t},\bx_{t}^{*})\} \\
& = \bbE[\bigl[U_{t}(\tilde{z}_{t}, \bx_{t})-U_{t}(z^{*}_{t}, \bx_{t})\bigr]\cdot \mathbf{1}\bigl\{\tilde{z}_{t} \neq z^{*}_{t}\bigr\}]+ \bbE[U_{t}(z^{*}_{t}, \bx_{t})-L_{t}(z^{*}_{t}, \bx_{t})] \\
& + 2 \cdot \bbP\{ \bigl[ f_{\theta_{i_{t},t}}(\bx_{t}) < L_{t}(z^{*}_{t}, \bx_{t})\bigr] \cup \bigl[ f_{\theta_{i_{t},t}}(\bx_{t}^{*}) > U_{t}(z^{*}_{t},\bx_{t}^{*})\bigr] \}
\end{split}
\end{equation*}
\normalsize
Then summing over $T$ we have:
\small
\begin{equation*}
\begin{split}
    \bbE[R_{T}] & \leq \sum_{t=1}^{T}\bbE[U_{t}(z^{*}_{t}, \bx_{t})-L_{t}(z^{*}_{t}, \bx_{t})] \\
    & + 2 \cdot \sum_{t=1}^{T}\bbP\{ \bigl[ f_{\theta_{i_{t},t}}(\bx_{t}) < L_{t}(z^{*}_{t}, \bx_{t})\bigr] \cup \bigl[ f_{\theta_{i_{t},t}}(\bx_{t}^{*}) > U_{t}(z^{*}_{t},\bx_{t}^{*})\bigr] \} \\
    & + \sum_{t=1}^{T}\bbE[\bigl[U_{t}(\tilde{z}_{t}, \bx_{t})-U_{t}(z^{*}_{t}, \bx_{t})\bigr]\cdot \mathbf{1}\bigl\{\tilde{z}_{t} \neq z^{*}_{t}\bigr\}] \\
    & \leq \sum_{t=1}^{T}\bbE[U_{t}(z^{*}_{t}, \bx_{t})-L_{t}(z^{*}_{t}, \bx_{t})] \\
    & + 2 \cdot \sum_{t=1}^{T}\bbP\{ \bigl[ f_{\theta_{i_{t},t}}(\bx_{t}) < L_{t}(z^{*}_{t}, \bx_{t})\bigr] \cup \bigl[ f_{\theta_{i_{t},t}}(A_{t}^{*}) > U_{t}(z^{*}_{t},\bx_{t}^{*})\bigr] \} \\
        & + \sum_{t=1}^{T}\bbE[\bigl[U_{t}(z^{*}_{t}, \bx_{t}^{*})-U_{t}(z^{*}_{t}, \bx_{t})\bigr]\cdot \mathbf{1}\bigl\{\tilde{z}_{t} \neq z^{*}_{t}\bigr\}]
\end{split}
\end{equation*}
\normalsize

\section{Proof of Lemma \ref{lemma:wrong_model}}
We can rewrite $\sum_{t=1}^{T}P(\tilde{z}_{t} \neq z^{*}_{t}|\mathcal{L}_{t}^{C})$ as follows:
\small
\begin{equation} \label{eq:wrong_model_decomposition}
\begin{split}
    & \sum_{t=1}^{T}P(\tilde{z}_{t} \neq z^{*}_{t}|\mathcal{L}_{t}^{C}) = \sum_{u \in \cU, c_{u,i} \in \cC_{u,T}} \sum_{t\in S_{u,c_{u,i}}} P(\tilde{z}_{t} \neq z^{*}_{t}|\mathcal{L}_{t}^{C}) \\ 
    & = \sum_{u \in \cU, c_{u,i} \in \cC_{u,T}} \sum_{t\in S_{u,c_{u,i}}} \sum_{k \in [K_{t}]} P(z^{*}_{u,c_{u,i}}=k) P(\tilde{z}_{t} \neq k|\mathcal{L}_{t}^{C},z^{*}_{u,c_{u,i}}=k) \\ 
    & = \sum_{u \in \cU, c_{u,i} \in \cC_{u,T}} \sum_{t\in S_{u,c_{u,i}}} \sum_{k \in [K_{t}]} \frac{n_{k,c_{u,i}-1}}{n_{c_{u,i}-1}} P(\tilde{z}_{t} \neq k|\mathcal{L}_{t}^{C},z^{*}_{u,c_{u,i}}=k) \\ 
\end{split}
\end{equation}
\normalsize
where the first equality is simply rewriting the summation over each stationary period of each user. Recall that $S_{u,c}$ denotes the stationary period after the $c$'th change point of user $u$. Note that $n_{t}$ denotes the total number of stationary periods among all users up to time $t$, and $n_{k,t}$ denotes the number of stationary periods whose bandit parameter equals to $\phi_{k}$ up to time $t$.

Let's denote the value of marginalized likelihood function $P(\cD^{u_{t}}_{t}|\tilde z_{t}=k, \cG)$ of model $k$ as $f_{k,t}(\cD^{u_{t}}_{t})$ for simplicity. Then by applying Eq. \ref{eq:posterior_predictive} and the inequality that $1 - \frac1x \leq \log x \leq x-1$ for all $x > 0$, we have:
\begin{align*}
    & P(\tilde{z}_{t} \neq k|\mathcal{L}_{t}^{C},z^{*}_{u,c_{u,i}}=k) = 1- P(\tilde{z}_{t} = k|\mathcal{L}_{t}^{C},z^{*}_{u,c_{u,i}}=k) \\
    & = 1-\frac{n_{k,c_{u,i}-1}f_{k,t-1}(\cD^{u}_{t-1})}{\sum_{k^{\prime}\in[K]}n_{k^{\prime},c_{u,i}-1}f_{k^{\prime},t-1}(\cD^{u}_{t-1})} \\
    & = 1-\frac{1}{1+\sum_{k^{\prime} \neq k}\frac{n_{k^{\prime},c_{u,i}-1}f_{k^{\prime},t-1}(\cD^{u}_{t-1})}{n_{k,c_{u,i}-1}f_{k,t-1}(\cD^{u}_{t-1})}} \leq \sum_{k^{\prime} \neq k}\frac{n_{k^{\prime},c_{u,i}-1}f_{k^{\prime},t-1}(\cD^{u}_{t-1})}{n_{k,c_{u,i}-1}f_{k,t-1}(\cD^{u}_{t-1})} \\
\end{align*}

Then look at the ratio between marginalized likelihood ratio:
\small
\begin{align*}
& \frac{f_{k^{\prime},t-1}(\cD^{u}_{t-1})}{f_{k,t-1}(\cD^{u}_{t-1})}=\prod_{(\bx_{i},r_{i}) \in \cD^{u}_{t-1}} \frac{\cN\big(r_{i}| \bx_{i}^{\mt} \mu_{k^{\prime},t-1}, \sigma^2 + \bx_{i}^\mt \Sigma_{k^{\prime},t-1}^{-1} \bx_{i}\big)}{\cN\big(r_{i}| \bx_{i}^{\mt} \mu_{k,t-1}, \sigma^2 + \bx_{i}^\mt \Sigma_{k,t-1}^{-1} \bx_{i}\big)} \\
& =\prod_{(\bx_{i},r_{i}) \in \cD^{u}_{t-1}} \frac{\sqrt{\bx_{i}^\mt \Sigma_{k,t-1}^{-1} \bx_{i} + \sigma^2}}{\sqrt{\bx_{i}^\mt \Sigma_{k^{\prime},t-1}^{-1} \bx_{i} + \sigma^2}} \\
& \cdot \exp{\bigl(\frac{1}{2}[\underbrace{\frac{(r_{i}-\bx_{i}^{\top}\mu_{k,t-1})^{2}}{\bx_{i}^\mt \Sigma_{k,t-1}^{-1} \bx_{i}+\sigma^2}}_{D_{k}}-\underbrace{\frac{(r_{i}-\bx_{i}^{\top}\mu_{k^{\prime},t-1})^{2}}{\bx_{i}^\mt \Sigma_{k^{\prime},t-1}^{-1} \bx_{i}+\sigma^2}}_{D_{k^{\prime}}}]\bigr)} \\
\end{align*}
\normalsize

Note that $\frac{\sqrt{\bx_{i}^\mt \Sigma_{k,t-1}^{-1} \bx_{i} + \sigma^2}}{\sqrt{\bx_{i}^\mt \Sigma_{k^{\prime},t-1}^{-1} \bx_{i} + \sigma^2}} \leq \sqrt{\frac{\bx_{i}^{\top}(\lambda I)^{-1}\bx_{i}+\sigma^{2}}{\sigma^{2}}} \leq \sqrt{\frac{1}{\sigma^{2}\lambda}+1}$. Then if $\sqrt{\frac{1}{\sigma^{2}\lambda}+1} \cdot \exp{\frac{1}{2}(D_{k}-D_{k^{\prime}})} < 1$ (which means $D_{k}-D_{k^{\prime}} < \log{\frac{\sigma^{2}\lambda}{\sigma^{2}\lambda + 1}}<0$), the ratio $\frac{f_{k^{\prime},t-1}(\cD^{u}_{t-1})}{f_{k,t-1}(\cD^{u}_{t-1})}$ can be shown to exponentially decrease as the size of $\cD^{u}_{t-1}$ grows.

Recall that ground truth model for observations $\cD^{u}_{t-1}$ is $k$, so $D_{k} \sim \chi^{2}(d=1)$ where $\chi^{2}(d=1)$ denotes a $\chi^{2}$ distribution with degree of freedom $d=1$. And similarly $D_{k^{\prime}} \sim \chi^{2}(d=1,\psi)$, which is a non-central $\chi^{2}$ distribution with degree of freedom $d=1$, and non-centrality parameter $\psi=\frac{(\bx_{i}^{\top}(\phi_{k}-\phi_{k^{\prime}}))^{2}}{\bx_{i}^\mt \Sigma_{k^{\prime},t-1}^{-1} \bx_{i}+\sigma^2} \geq \frac{\Delta^{2}}{\bx_{i}^\mt \Sigma_{k^{\prime},t-1}^{-1} \bx_{i}+\sigma^2}$. Intuitively, the non-centrality parameter $\psi$ controls how much these two distributions overlap with each other, and it depends on the gap $\Delta$ between the ground-truth bandit parameter projected on $\bx_{i}$ and the variance of the marginalized likelihood.

Denote the cumulative density function of non-central $\chi^{2}$ distribution as $F(\upsilon;d=1,\psi)$. Then with probability $F(\upsilon;d=1)$, $D_{k} \leq \upsilon$. Let's set $F(\upsilon;d=1)=1-\delta$, then $D_{k} \leq F^{-1}(1-\delta|d=1)$, with probability $1-\delta$, where $F^{-1}(\cdot|d)$ denotes inverse function of $F$.

Recall that we want $D_{k}-D_{k^{\prime}} < \log{\frac{\sigma^{2}\lambda}{\sigma^{2}\lambda + 1}}<0$, and to satisfy this, we need $D_{k^{\prime}} > F^{-1}(1-\delta|d=1) - \log{\frac{\sigma^{2}\lambda}{\sigma^{2}\lambda + 1}}$.
Similarly, using the cumulative density function, we know that this inequality hold with probability $1-F(F^{-1}(1-\delta|d=1) - \log{\frac{\sigma^{2}\lambda}{\sigma^{2}\lambda + 1}}|d=1,\psi)$. 



Define function $g(\psi;d,\upsilon)=F(\upsilon;\psi,d)$, and $g^{-1}(\cdot;d,\upsilon)$ as its inverse function. Then to make $1-F(F^{-1}(1-\delta|d=1) - \log{\frac{\sigma^{2}\lambda}{\sigma^{2}\lambda + 1}}|d=1,\psi) \geq 1-\delta$, we need $\psi \geq g^{-1}(\delta;1,F^{-1}(\delta;d=1)-\log{\frac{\sigma^{2}\lambda}{\sigma^{2}\lambda + 1}})$ and we denote this lower bound as $\psi_{L}$.

This result tells us that if the non-central parameter $\psi$ is greater than the constant $\psi_{L}$, then with probability $1-\delta$, the ratio $\frac{f_{k^{\prime},t-1}(\cD^{u}_{t-1})}{f_{k,t-1}(\cD^{u}_{t-1})}$ exponentially decreases to $0$ as the size of $\cD^{u}_{t-1}$ grows. And the decrease rate is denoted as $C_{2}=\sqrt{\frac{1}{\sigma^{2}\lambda}+1} \cdot \exp{\frac{1}{2}(D_{k}-D_{k^{\prime}})}<1$. This leads to a constant growth rate of regret within each stationary period $S_{u,c_{u,i}}$, e.g. $\lim_{|S_{u,c_{u,i}}| \rightarrow \infty} \sum_{i}^{|S_{u,c_{u,i}}|} C_{2}^{i} = \frac{C_{2}}{1-C_{2}}$.

\small
\begin{equation} \label{eq:wrongmodel_1}
\begin{split}
    &\sum_{t=1}^{T}P(\tilde{z}_{t} \neq z^{*}_{t}|\mathcal{L}_{t}^{C}) \\
    & = \sum_{u \in \cU, c_{u,i} \in \cC_{u,T}} \sum_{i}^{|S_{u,c_{u,i}}|} \sum_{k \in [K_{t}]} \frac{n_{k,c_{u,i}-1}}{n_{c_{u,i}-1}} (\sum_{k^{\prime} \neq k}\frac{n_{k^{\prime},c_{u,i-1}}}{n_{k,c_{u,i-1}}})  C_{2}^{i} \\
    & \leq \sum_{u \in \cU, c_{u,i} \in \cC_{u,T}} \sum_{i}^{|S_{u,c_{u,i}}|} K_{T} C_{2}^{i} = O(K_{T} \sum_{u \in \cU} \Gamma^{u}_{T}) \\
\end{split}
\end{equation}
\normalsize

On the other hand, at the time steps when $\psi\geq \psi_{L}$ does not hold, we suffer additional regret upper bounded by $2$ in each of these time steps. Recall that $\psi \geq \frac{\Delta^{2}}{\bx_{i}^\mt \Sigma_{k^{\prime},t-1}^{-1} \bx_{i}+\sigma^2} \geq \frac{\Delta^{2} / \sigma^{2}}{1/\lambda_{min}(\sigma^{2}\Sigma_{k^{\prime},t-1})+1}$. If $\lambda_{min}(\sigma^{2}\Sigma_{k^{\prime},t-1}) \geq \frac{\psi_{L}\sigma^{2}}{\Delta^{2}}$, the condition $\psi\geq \psi_{L}$ will hold. Therefore, we can upper bound the additional regret in terms of the total number of time steps that the minimum eigenvalue $\lambda_{min}(\sigma^{2}\Sigma_{k,t-1}) < \frac{\psi_{L}\sigma^{2}}{\Delta^{2}}$, for $k \in [K]$: $2\sum_{k \in [K]} \sum_{t \in \cI(k)} \textbf{1}\left\{\lambda_{min}(\sigma^{2}\Sigma_{k,t-1}) < \frac{\psi_{L}\sigma^{2}}{\gamma^{2}} \right\}$.

Our analysis follows a similar procedure as that of \cite{gentile2014online,pmlr-v70-gentile17a}. Borrowing the notation from \cite{pmlr-v70-gentile17a}, denote $A_{t}$ as a correlation matrix constructed through a series of rank-one updates using context vectors from $\left\{\cA_{t}\right\}_{t \in S}$, where $S$ denotes the set of time steps we performed model update. Note that the choice of which context vector to select from $\cA_{t}$ for $t\in S$ can be arbitrary. Then we denote the maximum number of updates it takes until $\lambda_{\min}(A_{t})$ is lower bounded by $\eta$ as 
\small $HD(\left\{C_{t}\right\}_{t \in S}, \eta) =\max\left\{t \in S: \exists \bx_{1} \in C_{1},...,\bx_{t} \in C_{t}: \lambda_{\min}(A_{t}) \leq \eta \right\}$ \normalsize, where $A_{t}=\sum_{u \in S: u \leq t}\bx_{u}\bx_{u}^{\top}$. Therefore, we obtain:
\small
\begin{equation} \label{eq:wrongmodel_2}
\begin{split}
& \sum_{k \in [K_{T}]} \sum_{t \in \cI(k)} \textbf{1}\left\{\lambda_{min}(\sigma^{2}\Sigma_{k,t-1}) < \frac{\psi_{L}\sigma^{2}}{\gamma^{2}} \right\} \\
& \leq \sum_{k \in [K_{T}]} HD\Big(\left\{\cA_{t}\right\}_{t \in  \cI(k)},\frac{\psi_{L}\sigma^{2}}{\gamma^{2}}\Big) \leq \sum_{k \in [K_{T}]} O\Big(\frac{\psi_{L}\sigma^{2}}{\gamma^{2}{\lambda^{'}}^{2}}\log{\frac{d}{\delta^{'}}}\Big)
\end{split}
\end{equation}
\normalsize
with probability at least $1-\delta^{'}$, and the second inequality is obtained by applying Lemma 1 of \cite{pmlr-v70-gentile17a}.

Then combining the regret upper bound from Eq. \eqref{eq:wrongmodel_1} and Eq. \eqref{eq:wrongmodel_2}, we have
\begin{align*}
    \sum_{t=1}^{T}P(\tilde{z}_{t} \neq z^{*}_{t}|\mathcal{L}_{t}^{C}) = O(K_{T} \bigl[(\sum_{u \in \cU} \Gamma^{u}_{T})+\frac{\psi_{L}\sigma^{2}}{\gamma^{2}{\lambda^{'}}^{2}}\log{\frac{d}{\delta^{'}}}\bigr])
\end{align*}

\end{document}